%% file: main.tex
\documentclass[nohyperref]{article}
\PassOptionsToPackage{dvipsnames}{xcolor}

\usepackage{microtype}
\usepackage{graphicx}
\usepackage{booktabs} %

\usepackage{hyperref}

\usepackage[accepted]{icml2022}

\usepackage{amsmath}
\usepackage{amssymb}
\usepackage{mathtools}
\usepackage{amsthm}

\usepackage[capitalize,noabbrev]{cleveref}

\usepackage[utf8]{inputenc} %
\usepackage[T1]{fontenc}    %
\usepackage{hyperref}       %
\usepackage{url}            %
\usepackage{booktabs}       %
\usepackage{amsfonts}       %
\usepackage{nicefrac}       %
\usepackage{microtype}      %
\input{math_commands.tex}

\usepackage{enumerate}
\usepackage{graphicx}
\usepackage[export]{adjustbox}
\usepackage{booktabs}
\usepackage{tabu}
\usepackage{tikz}
\usepackage{tikzsymbols}
\usetikzlibrary{arrows.meta}
\usetikzlibrary{shapes.misc, positioning}
\usetikzlibrary{shapes,arrows}
\usepackage{pgfplots}
\usepgfplotslibrary{groupplots}
\usepackage{multirow}
\usepackage{wrapfig}
\usepackage{subcaption}
\usepackage[toc,page]{appendix}
\usepackage[normalem]{ulem}
\usepackage{paralist}
\usepackage{cleveref}
\usepackage{makecell}
\usepackage{enumitem}
\usepackage{xspace}
\usepackage{array}
\usepackage{bbm}
\usepackage{enumitem}
\usepackage{pifont}
\usepackage{braket}

\DeclarePairedDelimiter\abs{\lvert}{\rvert}

\newcommand{\knnlm}{$k$NN-LM}
\newcommand{\knn}{$k$NN}
\newcommand{\kmeans}{$k$-means}
\newcommand{\ourmodel}{{\sc{RetoMaton}}}
\newcommand{\bourmodel}{{\sc{\textbf{RetoMaton}}}}

\newcommand{\adaptret}{{\sc{AdaptRet}}}
\newcommand{\badaptret}{{\sc{\textbf{AdaptRet}}}}
\newcommand{\wiki}{{\sc{Wikitext-103}}}
\newcommand{\mydot}{\raisebox{1pt} {\tikz\draw[black,fill=black] (0,0) circle (.3ex);}}
\definecolor{myblue}{HTML}{6C8EBF}
\definecolor{mygreen}{HTML}{82B366}
\definecolor{myred}{HTML}{b85450}
\DeclareRobustCommand\mycircle{\raisebox{-2pt} {\tikz[]{\node[shape=circle,draw=myred, line width=0.5mm,inner sep=1pt]{\phantom{\scriptsize{?}}};}}}
\DeclareRobustCommand\mycircleblue{\raisebox{-2pt} {\tikz[]{\node[shape=circle,draw=myblue, line width=0.5mm,inner sep=1pt]{\phantom{\scriptsize{?}}};}}}
\DeclareRobustCommand\mycirclegreen{\raisebox{-2pt} {\tikz[]{\node[shape=circle,draw=mygreen, line width=0.5mm,inner sep=1pt]{\phantom{\scriptsize{?}}};}}}
\DeclareRobustCommand\myroundrect{\raisebox{-2pt} {\tikz[]{\node[rounded rectangle,draw=myred, line width=0.5mm,inner sep=2pt,
](b){\phantom{\scriptsize{?}}};}}}

\newcommand{\myarrow}[1]{\tikz\draw[-{Triangle[width=6pt,length=6pt]},line width=2pt,fill=#1, draw=#1](0, 0) -- (0.35, 0);}
\newcommand{\redarrow}{\myarrow{myred}}
\newcommand{\greenarrow}{\myarrow{mygreen}}
\newcommand{\bluearrow}{\myarrow{myblue}}
\newcommand{\dashedarrow}{\tikz\draw[-{Triangle[width=4pt,length=4pt]},densely dashed,line width=1pt](0, 0) -- (0.35, 0);}
\newlength\myheight
\newlength\mydepth
\settototalheight\myheight{Xygp}
\newcommand*{\img}[1]{%
    \raisebox{-1.1\mydepth}{%
        \includegraphics[
        height=1.2\myheight,
        keepaspectratio,
        ]{#1}%
    }%
}
\newcommand{\circnum}[1]{\raisebox{.5pt}{\textcircled{\raisebox{-1pt} {#1}}}}
\renewcommand{\paragraph}[1]{\vspace{3pt}\noindent\textbf{\textit{#1}}}

\theoremstyle{plain}

\theoremstyle{definition}

\theoremstyle{remark}

\usepackage[textsize=tiny]{todonotes}

\newcommand{\uri}[1]{{\color{OliveGreen}Uri:[#1]}}
\newcommand{\gn}[1]{\textcolor{magenta}{[GN: #1]}}
\newcommand{\fx}[1]{\textcolor{red}{[Frank: #1]}}
\newcommand{\jh}[1]{\textcolor{brown}{[JH: #1]}}

\renewcommand{\uri}[1]{}
\renewcommand{\gn}[1]{}
\renewcommand{\fx}[1]{}
\renewcommand{\jh}[1]{}

\newcommand{\mytitle}{Neuro-Symbolic Language Modeling with Automaton-augmented Retrieval}
\icmltitlerunning{\mytitle{}}

\begin{document}

\twocolumn[
\icmltitle{\mytitle{}}

\begin{icmlauthorlist}
    \icmlauthor{Uri Alon}{lti}
    \icmlauthor{Frank F. Xu}{lti}
    \icmlauthor{Junxian He}{lti}
    \icmlauthor{Sudipta Sengupta}{amazonaws}
    \icmlauthor{Dan Roth}{amazon}
    \icmlauthor{Graham Neubig}{lti} \\
    \begin{tabular}{lcr} 
        \textsuperscript{1}Language Technologies Institute, Carnegie Mellon University &
    \textsuperscript{2}Amazon AWS &
    \textsuperscript{3}AWS AI Labs
    \end{tabular} \\
    \begin{tabular}{lr}
    \texttt{\{ualon,fangzhex,junxianh,gneubig\}@cs.cmu.edu} &
    \texttt{\{sudipta,drot\}@amazon.com}
    \end{tabular} 
\end{icmlauthorlist}

\icmlaffiliation{lti}{Language Technologies Institute, Carnegie Mellon University}
\icmlaffiliation{amazonaws}{Amazon AWS}
\icmlaffiliation{amazon}{Amazon}

\icmlcorrespondingauthor{Uri Alon}{ualon@cs.cmu.edu}

\icmlkeywords{Machine Learning, ICML}

\vskip 0.3in
]

\printAffiliationsAndNotice{}  %

\input{00_abstract.tex}
\input{01_intro.tex}
\input{02_background.tex}
\input{03_automaton.tex}
\input{04_setup.tex}

\input{05_experiments.tex}

\input{06_ablation.tex}

\input{07_qualitative.tex}

\input{08_related.tex}
\input{09_conclusion.tex}
\input{10_acks.tex}

\bibliography{biblio.bib}
\bibliographystyle{icml2022}

\newpage
\input{appendix_main.tex}
\end{document}

%% file: math_commands.tex
\usepackage{amsmath,amsfonts,bm}

\def\eqref#1{equation~\ref{#1}}

\def\1{\bm{1}}

\DeclareMathAlphabet{\mathsfit}{\encodingdefault}{\sfdefault}{m}{sl}
\SetMathAlphabet{\mathsfit}{bold}{\encodingdefault}{\sfdefault}{bx}{n}

%% file: 00_abstract.tex
\begin{abstract}
Retrieval-based language models (R-LM) model the probability of natural language text by combining a standard language model (LM) with examples retrieved from an external datastore at test time. 
While effective, a major bottleneck of using these models in practice is the computationally costly datastore search, 
which can be performed as frequently as every time step.
In this paper, we present \ourmodel{} -- retrieval automaton -- which approximates the datastore search, based on 
(1) saving \emph{pointers} between consecutive datastore entries, and
(2) clustering of entries into ``states''.
This effectively results in a \emph{weighted finite automaton} built on top of the datastore, instead of representing the datastore as a flat list.
The creation of the automaton is \emph{unsupervised}, and a \ourmodel{} can be constructed from any text collection: either the original training corpus or from another domain. 
Traversing this automaton at inference time, in parallel to the LM inference, reduces its perplexity by up to 1.85, or alternatively
saves up to 83\% of the nearest neighbor searches over \knnlm{} \cite{khandelwal2020generalization} without hurting perplexity. 
Our code and trained models are available at \url{https://github.com/neulab/retomaton} .
\end{abstract}

%% file: 01_intro.tex
\section{Introduction}
\label{sec:intro}
\input{intro_fig.tex}

Retrieval-based language models (R-LMs) have recently been shown to improve over standard neural models in a variety of tasks such as unconditional language modeling \cite{guu2018generating,he2020learning},
machine translation \cite{zhang2018guiding,gu2018search,khandelwal2021nearest}, 
question answering \cite{karpukhin2020dense,ram2021learning}, and code generation \cite{hayati2018retrieval, hashimoto2018retrieve}.
The key ingredient of R-LMs is their ability to utilize training examples at test time without having to rely on the information encoded in the model's weights only.

In these models, the retrieval component first searches for nearest neighbor examples in an external datastore; then, the base model references these examples during the prediction. 
This fusion of language models (LMs) and %
retrieval improves the base language model from several perspectives, including higher accuracy \cite{xu2021capturing}, domain adaptability \cite{jiang2021learning}, and reduced size \cite{borgeaud2021improving}. Further, the retrieved examples provide information regarding the provenance of the model's predictions, and retrieval allows for modifying the dataset without retraining the model. Nevertheless, the most critical bottleneck of these models is their \emph{frequent search} over the datastore, which hinders the use of R-LMs in practical settings.

\paragraph{$k$-Nearest Neighbors Language Model}
One prominent example of such a retrieval-based model is \knnlm{} \cite{grave2017unbounded,khandelwal2020generalization}, which predicts a token by linearly interpolating the base LM's output with a non-parametric nearest neighbor distribution. This distribution is constructed by searching for the $k$-nearest neighbors (\knn{}) in the datastore and  weighting them according to their distance to the current test context.
Notably, this $k$-nearest neighbor search is performed \emph{for every generated test token}, introducing severe inference overhead, since this search is significantly slower than the LM's standard ``forward pass''.

\paragraph{Our Approach: \bourmodel{}}
Our main insight is that 
\emph{retrieved neighbors at the current time step also hint at the neighbors that will be retrieved at future time steps},
 and can thus save repetitive searches later. Specifically, we construct a \emph{weighted finite automaton} (WFA) on top of an existing datastore, by keeping pointers between datastore entries and clustering similar entries, in a \emph{completely unsupervised} way.
This automaton allows sporadic, infrequent, \knn{} searches, and a much cheaper traversal of the automaton
at other time steps. We call our model \ourmodel{} -- retrieval automaton. \ourmodel{} is illustrated in \Cref{fig:intro}.

Concretely, applying \ourmodel{} to a strong \wiki{} LM using only the original training set allows for  saving 81\% of the \knn{} searches of \citet{khandelwal2020generalization} without hurting perplexity, or alternatively, reducing perplexity by 1.85 without saving searches. In both cases, we do not perform any additional training other than clustering. 
We also show that \ourmodel{} allows for effective domain adaptation, by simply constructing a \ourmodel{} for a different domain. When we construct \ourmodel{} on top of a \emph{fine-tuned} LM, we decrease the perplexity by more than 17\% 
compared to just fine-tuning.
Finally, we perform a thorough ablation study, separating the contributions of pointers and clustering, and analyzing the tradeoff between coarse- vs. fine-grained clustering.
We believe that 
these results suggest a promising direction for the neuro-symbolic synergy of neural models and symbolic automata.

%% file: intro_fig.tex
\begin{figure*}
\includegraphics[width=1\linewidth,keepaspectratio]{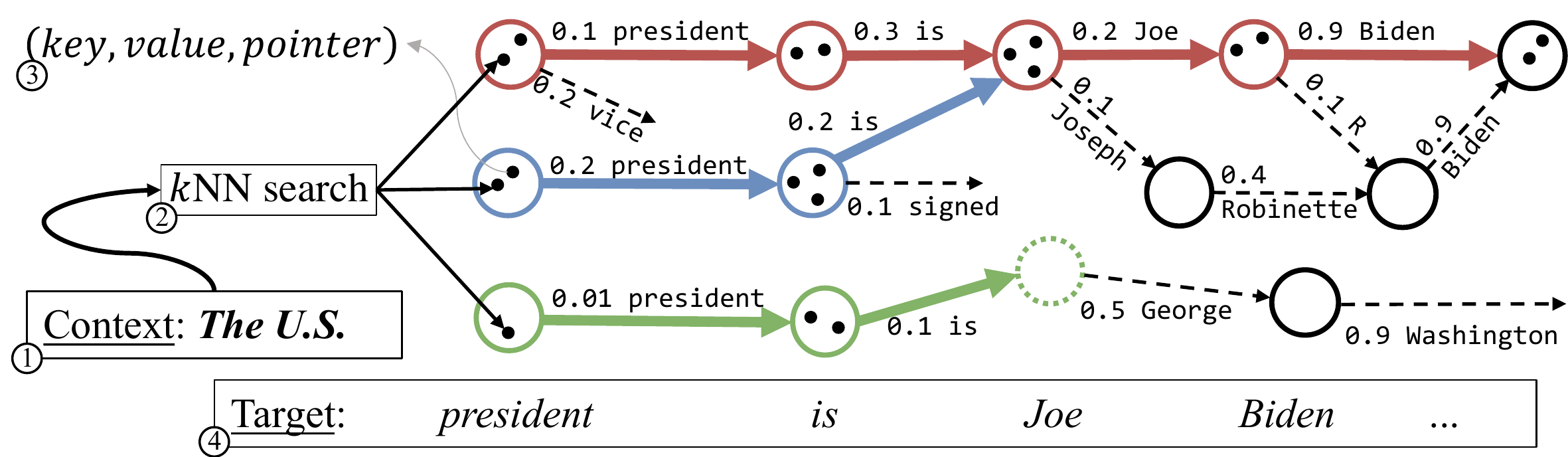}
\caption{An illustration of \ourmodel{}. 
Given a context \circnum{1} (``\emph{The U.S.}''), a $k$-nearest neighbor search \circnum{2} returns the nearest datastore entries.
Every datastore entry (\mydot{}\;  in the figure) is a 3-tuple of $\left(key,value,pointer\right)$ \circnum{3}, where the $key$ is the LM's hidden state and the $value$ is the target token as in \citet{khandelwal2020generalization}; the $pointer$ points to the datastore entry that appears next in the corpus. Close datastore entries are clustered together, and form an automaton state (\mycircle{},\mycircleblue{},\mycirclegreen{}). The pointers of the clustered entries form the state's possible transitions. At inference time, the model decodes \circnum{4} while performing multiple parallel traversals (\redarrow{}, \bluearrow{}, \greenarrow{}) on the automaton to find useful datastore entries, instead of performing a full \knn{} search. Dashed arrows (\dashedarrow) denote allowed automaton transitions that were not taken during the current decoding.
}
\label{fig:intro}
\end{figure*}

%% file: 02_background.tex
\section{Background: the \knnlm{} Model}
\label{sec:background}
\input{fig_zoomin.tex}

\knnlm{} \cite{khandelwal2020generalization} is a language model that estimates the probability of the next token by interpolating an already-trained base LM with a \knn{} distribution. The \knn{} distribution is provided by searching for the $k$-nearest neighbors in an external datastore and weighting them according to their negative distance.  

\paragraph{Datastore Creation} 
Given a context sequence of tokens $c^{\left(t\right)}=\left(w^{\left(1\right)},...,w^{\left(t-1\right)}\right)$, the base LM estimates $p_{LM}\left(w \mid c^{\left(t\right)}\right)$, the distribution over the target token $w$.
Let $f$ be the function that maps a context $c$ to a fixed-length vector using an already-trained LM. For example, $f\left(c\right)$ can be the output of a Transformer LM's last self-attention layer. 
\knnlm{} constructs the datastore using a single forward pass over a text collection, which can include the training set that the base LM was trained on, or not. Given the $i$-th example $\left(c_i,w_i\right)$ in the text collection $\mathcal{D}$, the key-value pair $\left(f\left(c_i\right),w_i\right)$ is defined such that $f\left(c_i\right)$ is the \emph{key}, and $w_i$ is the \emph{value}. The datastore $\left(\mathcal{K},\mathcal{V}\right)$ is the set of all pairs constructed from the examples in the text collection $\mathcal{D}$:
\begin{align}
\left(\mathcal{K},\mathcal{V}\right) = 
\set{
\left(f\left(c_i\right),w_i\right) \mid \left(c_i,w_i\right)\in \mathcal{D}
}
\end{align}
\paragraph{Inference} At test time, given a test context $c$, the model queries the datastore $\left(\mathcal{K},\mathcal{V}\right)$ to retrieve the $k$-nearest neighbors $\mathcal{N}$ of $f\left(c\right)$, according to a distance function $dist$ between $f\left(c\right)$ and every $f\left(c_i\right)$ in the datastore. 
$dist$ is typically the squared $\ell_2$ distance.
These nearest neighbor pairs form a distribution over the target token $w$, where the probability of every vocabulary item is computed proportionally to the exponent of its negative distance, while summing over all its occurrences in retrieved values of $\mathcal{N}$:
\begin{align}
& p_{\text{\knn{}}}\left(w \mid c\right)  \propto \nonumber \\
&  \sum_{\left(f\left(c_i\right),w_i\right)\in \mathcal{N}} \mathbbm{1}_{w=w_i} \exp\left(-dist\left(f\left(c\right),f\left(c_i\right)\right)\right)
\label{eq:knnscore}
\end{align}
The \knnlm{}'s next token probability is then the interpolation of $p_{LM}$ and $p_{\text{\knn{}}}$, using a scalar hyperparameter $\lambda$:
\begin{align}
\!\!\!\!
p\left(w \mid c\right)=
\lambda  p_{k\text{NN}}\left(w\mid c\right)+\left(1-\lambda\right)p_{LM}\left(w\mid c\right) \label{eq:interpolate} 
\end{align}
Notice that the $k$-nearest neighbor search is performed \emph{for every target token}. This introduces severe overhead, since the search is significantly slower than the LM's standard forward pass. In the next section, we show how we can avoid the search in the majority of the time steps.

%% file: fig_zoomin.tex
\begin{figure*}
\begin{center}
\includegraphics[width=0.8\linewidth,keepaspectratio]{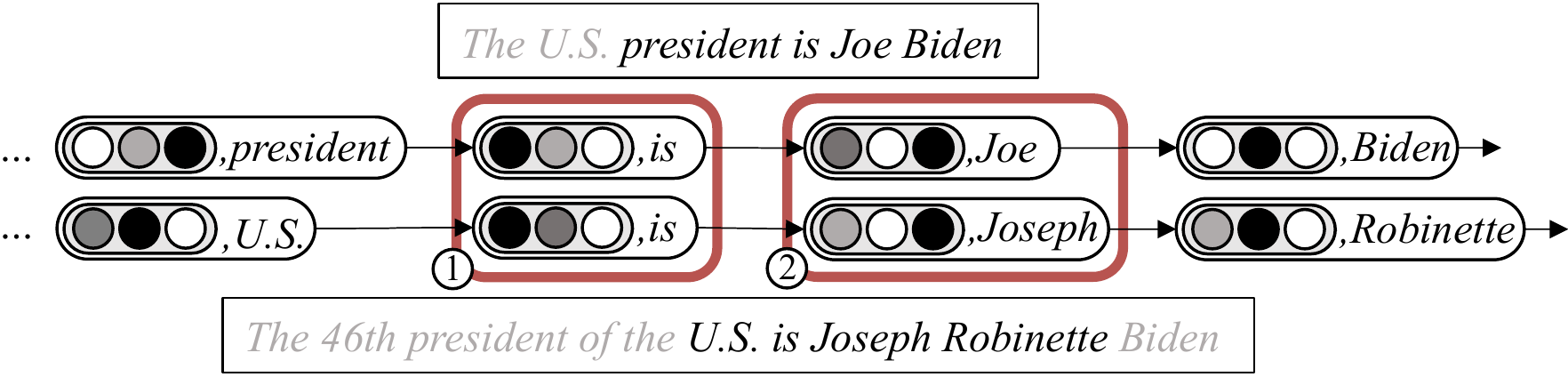}
\end{center}
\caption{
An illustration of our automaton creation process. The text $\mathcal{D}$ contains two sentences: 
\begin{inparaenum}[(i)]
	\item ``\emph{The U.S. president is Joe Biden}''; and
	\item ``\emph{The 46th president of the U.S. is Joseph Robinette Biden}''
\end{inparaenum}.
Each datastore entry is a pair of a key vector encoding the prefix and a value representing the next token, such as \img{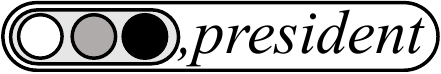}. We save a pointer from every entry to its successor in the text, and we cluster close key vectors to form automaton states (\circnum{1} and \circnum{2}) that share pointers. 
}
\label{fig:creation}
\end{figure*}

%% file: 03_automaton.tex
\section{Building Automata from Datastores}
\label{sec:automata}
To save \knn{} searches, we build a \emph{weighted finite automaton} (WFA) on top of the  \knnlm{} datastore.  Then, we traverse the automaton to estimate the next nearest neighbors.
We will demonstrate the automaton creation process using \Cref{fig:creation} as a running example.

\subsection{Definitions}
\label{subsec:def}
Given a finite vocabulary $\Sigma$ and the set $\Sigma^{*}$ of finite sequences over $\Sigma$, a trained autoregressive LM defines a distribution $p_{LM}$ over $\Sigma$ for any given context $c\in \Sigma^{*}$.
Given such an LM having $f:\Sigma^{*}\rightarrow \mathbb{R}^d$ and a text collection $\mathcal{D}$, we can create a datastore $\left(\mathcal{K},\mathcal{V}\right)$ as detailed in \Cref{sec:background}. 
As shown in \Cref{fig:creation}, this results in a set of key-value entries such as \img{entry-crop.pdf}, where the key is the LM encoding of every prefix in $\mathcal{D}$, and the value is the following token.

\paragraph{Weighted Finite Automaton} Our automaton is a tuple $A=\left<Q,\Sigma,q_0,\delta,\phi\right>$ such that $Q$ is a finite set of states, $\Sigma$ is the LM's vocabulary, 
$q_0 \in Q$ is the initial state,
$\delta : Q \times \Sigma \rightarrow \mathbb{P}\left(Q\right)$ is a transition function where $\mathbb{P}\left(Q\right)$ denotes the power set of $Q$, and $\phi: Q \times \mathbb{R}^{d} \times \Sigma \rightarrow \mathbb{R}$ is a transition weight function. 
Unlike standard WFAs, the transition weights are \emph{dynamic}: notice that $\phi$ depends also on a vector in $\mathbb{R}^{d}$, in addition to a previous state and an input token.
Also, note that $\delta$ is defined such that we transition into a \emph{set} of states.

\subsection{Constructing the Automaton}
\label{subsec:constructing}
\paragraph{Pointers}
Our main insight is that during the creation of the datastore, we can keep  a \emph{pointer} from every datastore entry to the entry that appears next in the text $\mathcal{D}$.

Imagine that at time $t$, the test context is $c^{\left(t\right)}$~$=$~$\left(w^{\left(1\right)},...,w^{\left(t-1\right)}\right)$, and the model  retrieves a datastore entry $\left(f\left(c_i\right),w_i\right)$. If after the interpolation with the base LM (\Cref{eq:interpolate}), the model's generated token $w^{\left(t\right)}$ (by sampling or argmax) is equal to $w_i$, then the entry $\left(f\left(c_{i+1}\right),w_{i+1}\right)$ is likely to be a useful entry at time $t+1$, because $c_i$ was a near neighbor of $c^{\left(t\right)}$, and both these contexts were followed by the same token $w_i=w^{\left(t\right)}$. That is, our underlying assumption can be formulated as:
\begin{equation}
	f\left(c_i\right) \approx f\left(c^{\left(t\right)}\right) \implies f\left(c_i\cdot w\right) \approx f\left(c^{\left(t\right)}\cdot w\right)
\end{equation}
where $\approx$ denotes vector similarity,
and $c^{\left(t\right)}\cdot w$ is the continuation of the context $c^{\left(t\right)}$ using the token $w$. %

Thus, given the $i$-th example $\left(c_i,w_i\right)\in \mathcal{D}$, instead of keeping only the key-value pair $\left(f\left(c_i\right),w_i\right)$ as in \citet{khandelwal2020generalization}, we save every datastore entry as $\left(f\left(c_i\right),w_i,p_i\right)\in \left(\mathcal{K},\mathcal{V},\mathcal{P}\right)$, where $p_i$ is a pointer to the next entry, or the next entry's index in the datastore. 
This is illustrated as arrows in \Cref{fig:creation}, where every entry has a pointer to the entry that followed it in the text $\mathcal{D}$. For example, the entry \img{entry-crop.pdf} points to \img{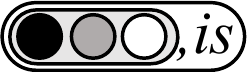}.

\paragraph{States}
If every entry had only a single outgoing pointer, the automaton would only capture n-grams 
that appeared in the text \emph{verbatim}, and thus will not be able to generalize to unseen phrases. For example in \Cref{fig:creation}, the sequence ``\emph{The 46th president of the U.S. is Joe Biden}'' would \emph{not} be captured, because the prefix ``\emph{The 46th president of the U.S.}'' appeared in one sentence, and the suffix ``\emph{Joe Biden}'' appeared in another sentence.
Thus, to allow entries to \emph{share} their pointers, we cluster entries having close keys into an automaton \emph{state}. A state includes all entries in the cluster, and the entries' pointers are the state's allowed outgoing  transitions. 
For example, in \Cref{fig:creation}, the entry \img{entry_is_cropped.pdf} is clustered with another entry having the same value, \img{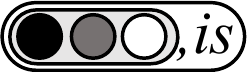} (surrounded by \myroundrect{} and marked as \circnum{1}). Furthermore, we can also cluster similar contexts that do \emph{not} have the same value, such as cluster \circnum{2} containing \img{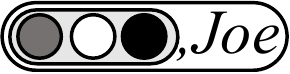} and \img{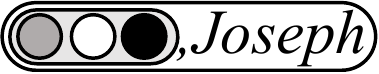}, which allows capturing the phrase ``\emph{Jospeh Biden}''. %

This step can be performed using any clustering algorithm such as $k$-means. In this case, the choice of $k_{\mathrm{\mathrm{clust}}}$ 
determines the final number of states $\abs{Q}=k_{\mathrm{clust}}$ 
(ignoring the initial state $q_0$). 
We experiment with different values of $k_{\mathrm{clust}}$ and clustering algorithms in \Cref{sec:analysis}.

\paragraph{Transition States}
Given a source state $q \in Q$ and an input token $w \in \Sigma$, we define the set of allowable next states as the clusters of entries that are pointed to by a datastore entry in $q$ having the value $w$. 
In other words, we follow pointers of datastore entries from $q$ whose value is $w$, and take the resulting entries' clusters.
Formally, let $\pi: \left(\mathcal{K},\mathcal{V},\mathcal{P}\right) \rightarrow Q$ be a function that maps every datastore entry into its containing state, and the function $\pi^{-1} :Q\rightarrow \mathbb{P}\left({\left(\mathcal{K},\mathcal{V},\mathcal{P}\right)}\right)$, %
which maps every state into the \emph{set} of datastore entries contained in it. Let $\rho: \mathcal{P}\rightarrow \left(\mathcal{K},\mathcal{V},\mathcal{P}\right)$, be the ``dereferencing'' operator, which returns a datastore entry given a pointer that points to it.
We can now define the allowed transitions as: 
\begin{equation}
	\medmuskip=2mu   %
	\thickmuskip=3mu %
	\renewcommand\arraystretch{1.5}
	\!\delta\left(q,w\right) = \set{\pi\left(\rho\left(p_i\right)\right) \mid \left(\cdot,w_i,p_i\right) \in \pi^{-1}\left(q\right), w_i=w} 
\end{equation}
This is illustrated in \Cref{fig:creation} as the outgoing pointers of cluster \circnum{2}, which allow transitioning from \circnum{2} to different states given the input tokens $w=$``\emph{Joe}'' or $w=$``\emph{Joseph}''.
This results in a (currently non-weighted) finite-state automaton, whose nodes are clusters of datastore entries, and whose edges represent the successiveness of entries in $\mathcal{D}$, where successiveness is \emph{shared} among clustered entries.

\subsection{Traversal of the Automaton}
\label{subsec:traversal}
At test time, given a test context $c^{\left(t\right)}$, we traverse the automaton while visiting \emph{multiple} states at every time step, marked in \Cref{fig:intro} as \redarrow{}, \bluearrow{}, \greenarrow{}. 
A traversal begins %
with a full \knn{} search of the datastore to retrieve the $k$-nearest neighbors $\mathcal{N}^{\left(t\right)}$
of $f\left(c^{\left(t\right)}\right)$.
The initial traversal states $\mathcal{S}^{\left(t\right)}\subseteq{Q}$ are the union of the states to which these $k$-nearest neighbors belong: $\mathcal{S}^{\left(t\right)}=\bigcup_{e \in \mathcal{N}^{\left(t\right)}}\pi\left(e\right)$.
\footnote{Formally, we start every traversal from the initial state $q_0$, perform a \knn{} search to retrieve $\mathcal{N}^{\left( t\right)}$, and then make an $\epsilon$-transition (transitioning without consuming an input token) into $\mathcal{S}^{\left(t\right)}$.
}

In the next time step $t+1$, given a token $w^{\left(t\right)}\in\Sigma$ that was generated (by argmax or sampling) by the model at time $t$, we compute the union of all valid transitions from states $q \in \mathcal{S}^{\left(t\right)}$. We define 
$\hat{\delta}:\mathbb{P}\left(Q\right)\times \Sigma\rightarrow \mathbb{P}\left(Q\right)$ as follows: 
\begin{align}
	\hat{\delta}\left(\mathcal{S}, w\right) =\bigcup_{q\in \mathcal{S}}\delta\left(q,w\right)
\end{align}
The decision of whether to continue traversing or start a new traversal can be made in several ways.
We experimented with several alternatives, but found the most intuitive way to simply be whether the number of new states is greater than or equal to a threshold $\tau$. That is, we continue traversing if $\abs{\hat{\delta}\left(\mathcal{S}^{\left(t\right)}, w^{\left(t\right)}\right)}\geq\tau$, or start a new traversal otherwise. 

When we \emph{continue traversing}, we take the new states 
as our states in the next step. When we \emph{start a new traversal}, we perform a new \knn{} search resulting in $\mathcal{N}^{\left({t+1}\right)}$, but also include the remaining states we obtained so far:
\begin{align}
	&\mathcal{S}^{\left(t+1\right)}=  \label{eq:restart} \\
	& \begin{cases}
		\hat{\delta}\left(\mathcal{S}^{\left(t\right)}, w^{\left(t\right)}\right) &  \abs{\hat{\delta}\left(\mathcal{S}^{\left(t\right)}, w^{\left(t\right)}\right)}\geq\tau \\
		\hat{\delta}\left(\mathcal{S}^{\left(t\right)}, w^{\left(t\right)}\right) \cup \!\!\!\!\bigcup\limits_{e\in \mathcal{N}^{\left(t+1\right)}}\!\!\!\!\!\!\pi\left(e\right)
		& \mathrm{otherwise}
	\end{cases} \nonumber
\end{align} 
Varying $\tau$ allows us to control the trade-off between higher accuracy with frequent traversal restarts, and thus frequent \knn{} searches (high $\tau$), versus lower accuracy  with rare \knn{} searches, which saves time (low $\tau$). For additional intuition for $\tau$, see \Cref{app:tau}.

\paragraph{Transition Weights}
Given a set of states $\mathcal{S}$, a test context $c$, and an input token $w \in \Sigma$, we define the transition weight from every $q\in \mathcal{S}$ similarly to 
\Cref{eq:knnscore}, except that we sum exponents of negative distances between $f\left(c\right)$ and all entries whose value is $w$ that are contained the state $q$;
then, we normalize across all states in $\mathcal{S}^{\left(t\right)}$:
\begin{flalign}
& \phi\left(q, c, w\right) = \!\!\!\!\!\!\!\sum_{\left(k_i,w_i,\cdot\right)\in\pi^{-1}\left(q\right)} \!\!\!\!\!\!\! \mathbbm{1}_{w=w_i} \exp\left(-dist\left(f\left(c\right),k_i\right)\right) \!\!\!\!&  \\
& p_{\text{auto}}\left(w \mid c, \mathcal{S}\right)  \propto
  \sum\nolimits_{q\in \mathcal{S}} \phi\left(q,c,w\right) & \label{eq:pauto}
\end{flalign}
Finally, we interpolate $p_{\text{auto}}$ with $p_{LM}$:
\begin{flalign}
&p\left(w \mid c,\mathcal{S}\right) =
\lambda  p_{\text{auto}}\left(w\!\mid\! c,\mathcal{S}\right)+\left(1\!-\!\lambda\right)p_{LM}\left(w\mid c\right) \!\!\!\!\!& \label{eq:interpolateretomaton}
\end{flalign} 

%% file: 04_setup.tex
\section{Experimental Setup}
\label{sec:setup}
We evaluate \ourmodel{} in two different settings: 
\begin{inparaenum}[(i)]
\item standard autoregressive language modeling, where the datastore is constructed from the same training corpus that the base LM was trained on (``in-training datastore''); and
\item domain adaptation, where the datastore is constructed from a \emph{different} domain than the base LM was trained on.
\end{inparaenum}
Our experiments are fully reproducible and our code is available at \url{https://github.com/neulab/retomaton} .

\paragraph{Implementation}
We base our experiments on the original \knnlm{} implementation that uses the FAISS \cite{johnson2019billion} library to perform \knn{} search. 
We also use FAISS for the one-time $k$-means clustering.

\paragraph{Hyperparameters} We used the same settings as the baseline implementations without any special tuning of our model, and always matched the settings to conduct a fair evaluation. 
We saved half precision (fp16) datastore keys as \citet{he2021efficient}.
For \wiki{}, which creates a datastore of 103M entries, we use \kmeans{} clustering with $k_{\mathrm{clust}}$$=$1M. For Law-MT, which creates a datastore of 19M entries, we use $k_{\mathrm{clust}}$$=$200K, which maintains an average cluster size of $\sim$100 in both datasets.
We analyze the difference between clustering algorithms and the number of clusters  in \Cref{sec:analysis}. Additional details are provided in \Cref{app:hyperparams}.

\paragraph{Metrics} 
The main metric that we focus on is perplexity with respect to the \emph{fraction of saved searches} (FoSS). 
In our preliminary experiments, we found that
measuring wall-clock time is difficult to reproduce, is brittle to %
temporary hardware and system load, and is affected by the specific \knn{} retrieval library.
Contrarily, FoSS does not depend on hardware and engineering factors and is thus completely reproducible. 
In \Cref{app:foss}, we empirically analyze  
 FoSS with respect to the saved wall-clock time and we show that they are identical up to an additive constant that depends on the hardware and the specific settings. Thus, FoSS serves as a good proxy to the saved wall-clock time. 

We control 
FoSS in \ourmodel{} by running with different values of the $\tau \in [1, \infty )$ threshold, as detailed in \Cref{subsec:traversal} and \Cref{app:tau}. Higher $\tau$ results in frequent restarts and thus lower FoSS. 

\subsection{In-Domain Datastore} 
\paragraph{Data}
Following \citet{khandelwal2020generalization}, we use
\wiki{} \cite{merity2016pointer}, which is a standard benchmark for autoregressive language modeling, having 103M/250K/250K tokens from Wikipedia in its training/validation/test sets, respectively. 

\paragraph{Model} We use a Transformer \cite{vaswani2017attention} as our base LM, trained by \citet{khandelwal2020generalization} following the architecture and settings of \citet{baevski2018adaptive} with adaptive inputs, adaptive softmax \cite{joulin2017efficient}, and a large 267K word-level vocabulary. This base LM consists of 16 layers, each with 16 self-attention heads, 1024-dimensional hidden states, 4096-dimensional feedforward layers, amounting to 247M parameters.

\subsection{Domain Adaptation}
\label{subsec:setup_domain_adapt}
\paragraph{Data} Following \citet{he2021efficient}, we use the English part of Law-MT, which is an English-German translation dataset for the law domain, released by \citet{koehn2017six} and resplit by \citet{aharoni2020unsupervised}. The training set consists of 19M tokens, which we use to build a datastore and an automaton (\Cref{subsec:domainadapt}), or fine-tune the base LM and then build a datastore and automaton~(\Cref{subsec:finetune}).

\paragraph{Model} Following \citet{he2021efficient}, as our base LM we use a 12-layer, 1536-dimensional transformer with a vocabulary of 42K subword units \cite{sennrich2016neural}, amounting to 656M parameters and trained by \citet{ng2019facebook} on WMT News Crawl \cite{barrault-etal-2019-findings}. 

\subsection{Baselines}
\label{subsec:baselines}
\paragraph{\knnlm{}} We compare \ourmodel{} to  \knnlm{} using the original code and hyperparameters of \citet{khandelwal2020generalization}. 
The only difference in hyperparameters  is that following \citet{he2021efficient}, we use the faster approximate \knn{} distances provided by FAISS, rather than recomputing them, in our model and in the baselines. %
We vary the FoSS in \knnlm{} by uniformly
selecting a certain fraction of 
time steps in which we skip the \knn{} search and use only the base LM ($p_{LM}$),
identically to the ``Random'' baseline in \citet{he2021efficient}. That is,  \knnlm with FoSS$=$0 is the standard \knnlm, and \knnlm{} with FoSS$=$1.0 is the base LM.

\paragraph{\badaptret{}} We also compare \ourmodel{} to a retrieval-saving approach that is  orthogonal to ours. \adaptret{} is the Adaptive Retrieval approach of \citet{he2021efficient}, which trains a light MLP to predict at each time step whether 
the base LM is ``confident enough'' to use its output only,
 or should it perform a \knn{} search.
The main conceptual difference between \ourmodel{} and \adaptret{} %
is that \ourmodel{} skips \knn{} searches but still computes the interpolation with the non-parametric distribution $p_{\text{auto}}$ for \emph{every} token, using the automaton (\Cref{eq:interpolateretomaton}). In contrast, when \adaptret{} skips a \knn{} search, it also skips the  interpolation with the $p_{k\mathrm{NN}}$ distribution of \Cref{eq:interpolate} entirely, and backs-off to rely on the base LM solely.
For a detailed discussion of the conceptual differences between \ourmodel{} and \adaptret{}, see \Cref{subsec:comparehe}.

%% file: 05_experiments.tex
\section{Results}
\label{sec:experiments}
\input{fig_wiki.tex}
\subsection{In-Domain Datastore}
\label{subsec:intraining}
We experiment with creating a datastore and an automaton from the same data that the base LM was trained on. 

\Cref{fig:wiki} shows how \ourmodel{} reduces the perplexity on \wiki{} across different FoSS rates.
Specifically, \ourmodel{} saves 81\% of the \knn{} searches while matching the perplexity of \knnlm{}. 
If we do not perform clustering (``\emph{w/o clustering}''), we can still save more than 60\% of the \knn{} searches while matching \knnlm{}.
Compared to \adaptret{}, \ourmodel{} saves 60\% of the searches while matching the best perplexity of \adaptret{}. 

Surprisingly, even when we do not attempt to save any searches (FoSS$=$0), \ourmodel{} reduces perplexity from 16.65 (\knnlm) and 16.35 (\adaptret{}) to 16.08. 
The explanation for this is that even when \ourmodel{} performs \knn{} search on \emph{every} step, as \knnlm{}, it includes the pointers from the previous time step (\Cref{eq:restart}), which are more likely to be correct than the general retrieved nearest neighbors. Some neighbors may be included twice -- both as retrieved \knn{}s, and as pointers from the previous time step; this case is equivalent to increasing the weight of a subset of the retrieved \knn{}s, which are more likely to be correct.

\input{fig_law.tex}

\subsection{Domain Adaptation}
\label{subsec:domainadapt}
We also experiment with domain adaptation, where the base LM was trained on newspapers, and the models are tested on law documents, using datastore and automaton that were constructed from law data as well, as detailed in \Cref{subsec:setup_domain_adapt}.

\Cref{fig:law} shows how \ourmodel{} reduces the perplexity on Law-MT from 12.34 (\knnlm{}) and 12.01 (\adaptret) to 10.49 when search is performed every step (FoSS$=$0). As we increase the fraction of saved searches, \ourmodel{} shows a very	 gentle ascent in perplexity, while the perplexity of \knnlm{} increases exponentially. 

The high perplexity of the \adaptret{} baseline is caused by the high perplexity of the base LM (106.56): 
in time steps where \adaptret{} does not perform search, its output is identical to the base LM's probability. That is, in domain adaptation, where the base LM performs poorly, interpolating it with the $p_{k\text{NN}}$ distribution (\Cref{eq:interpolate}) is \emph{crucial}. Thus, approximating the $p_{k\text{NN}}$ distribution using an automaton is much more effective than pruning it using \adaptret{}, while \knn{} searches are saved in both cases.

It is also interesting to notice the difference between datasets. In particular, we find that \ourmodel{} provides a stronger effect in Law-MT, reflected in the very gentle ascent in \Cref{fig:law}, over its effect in \wiki{}. 
We believe that one major reason is n-gram repetitiveness between the training  and the validation sets.
As shown in \Cref{fig:ngram_types,fig:ngram_occ}, there is much higher repetitiveness of n-grams in Law-MT over \wiki{}. For example, 21\% of the 5-grams in the validation set of \wiki{} were seen in the  training data; in contrast, in Law-MT -- 62\% of the 5-grams in the validation set were seen during training.

\subsection{Improving Fine-Tuning}
\label{subsec:finetune}
\input{fig_fintuned.tex}
In \Cref{subsec:domainadapt} we used \ourmodel{} to domain-adapt a base LM that was trained on a different domain; 
however, can \ourmodel{} improve a \emph{fine-tuned} base LM?

We fine-tuned the base LM of \Cref{subsec:domainadapt} on the Law-MT training set, and recreated a datastore and an automaton using the fine-tuned model.
As \Cref{fig:law-finetuned} shows, while \knnlm{} and \adaptret{} reduce the perplexity compared to the fine-tuned model from 8.61 to 7.93 and 7.81, respectively, \ourmodel{} further reduces perplexity to 7.10, which is a relative reduction of more than 17.5\%. This shows how \ourmodel{} \emph{can strengthen even fine-tuned models}.

%% file: fig_wiki.tex
\definecolor{ao}{rgb}{0.0, 0.5, 0.0}
\begin{figure}[t]
	\begin{tikzpicture}[scale=1]
	\begin{axis}[
		xlabel={FoSS (fraction of saved searches)},
		ylabel={Perplexity},
		ylabel near ticks,
        legend style={at={(0,1)},anchor=north west,mark size=2pt,
        	}, %
        legend cell align={left},
        xmin=0.0, xmax=1.0,
        ymin=15.95, ymax=18.95,
        xtick={0.1,0.2,...,1.0},
        ytick={0,1,2,...,19},
        ylabel shift={-5pt},        
        grid = major,
        major grid style={dotted,gray},
    ]

\addplot[color=red, solid, mark options={solid, fill=red, draw=black}, mark=triangle*, line width=1pt, mark size=3pt, visualization depends on=\thisrow{alignment} \as \alignment, nodes near coords, point meta=explicit symbolic,
    every node near coord/.style={anchor=\alignment, font=\footnotesize}] 
table [meta index=2]  {
x   y       label   alignment
0	16.65	16.65		0
0.1	16.84	{}		-0
0.3	17.21	17.21		-20
0.5	17.60	17.60		-20
0.7 18.02	18.02		160
0.9 18.44	18.44		-20
1.0	18.66	18.66		-20
	};
    \addlegendentry{\knnlm{}}

\addplot[color=blue, mark options={solid, fill=blue, draw=black}, mark=square*, line width=1pt, mark size=3pt, visualization depends on=\thisrow{alignment} \as \alignment, nodes near coords, point meta=explicit symbolic,
    every node near coord/.style={anchor=\alignment, font=\footnotesize}] 
table [meta index=2]  {
x   y       label   alignment
0	16.35	16.35		-10
0.1	16.36	{}		-90
0.3	16.47	16.47		-20
0.5	16.67	16.67		-20
0.7 16.99	16.99		-20
0.9 17.59	17.59		-20
1.0	18.66	{}		-66
	};
    \addlegendentry{\adaptret{}}

\addplot[color=ao, mark options={solid, fill=ao, draw=black}, mark=*, line width=1pt, mark size=3pt, visualization depends on=\thisrow{alignment} \as \alignment, nodes near coords, point meta=explicit symbolic,
    every node near coord/.style={anchor=\alignment, font=\small
    }] 
table [meta index=2]  {
x   y       label   alignment
0.0	16.08	\textbf{16.08}	-10
0.26	16.16	16.16	90
0.36	16.22	{}	-20
0.45	16.27	16.27	90
0.52	16.31	{}	0
0.61	16.37	16.37	90
0.72	16.47	{}	0
0.81	16.65	16.65	-180
	};
    \addlegendentry{\ourmodel{} (this work)}    
  
\addplot[color=gray!70, mark options={solid, fill=white, draw=black}, mark=otimes, line width=1pt, mark size=3pt, visualization depends on=\thisrow{alignment} \as \alignment, nodes near coords, point meta=explicit symbolic,
    every node near coord/.style={anchor=\alignment, font=\small
    }] 
table [meta index=2]  {
x   y       label   alignment
0	16.12	16.12	-35
0.20	16.18	{}	0
0.30	16.23	{}	0
0.37	16.29	{}	0
0.43	16.35	{}	0
0.50	16.43	{}	0
0.55	16.50	{}	0
0.59	16.57	{}	0
0.68	16.85	{}	0
	};
    \addlegendentry{\;\;w/o clustering}    
    
	\end{axis}
\end{tikzpicture}
\caption{Experiments on \wiki{}, where the datastore is created from the same training set that the base LM was trained on. \ourmodel{} reduces perplexity across all FoSS values, and even reduces perplexity when FoSS$=$0.}
\label{fig:wiki}
\end{figure}
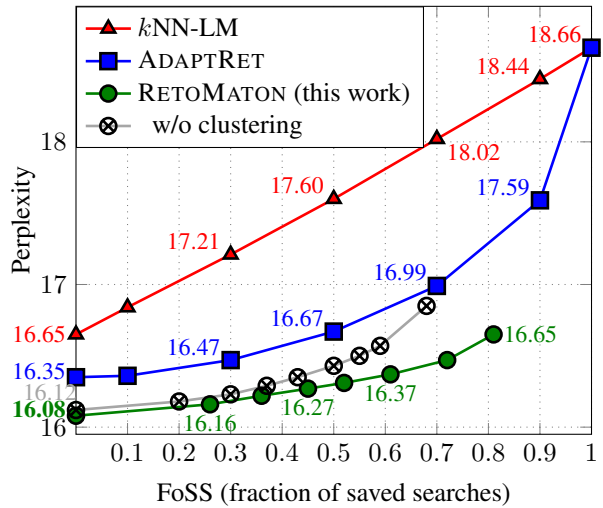 

%% file: fig_law.tex
\begin{figure}[t]
\begin{tikzpicture}[scale=1]
	\begin{axis}[
		xlabel={FoSS (fraction of saved searches)},
		ylabel={Perplexity},
		ylabel near ticks,
        legend style={at={(0,1)},anchor=north west,mark size=2pt,
        	}, %
        legend cell align={left},
        xmin=-0.1, xmax=1.0,
        ymin=0.0, ymax=110,
        xtick={0.0, 0.1,0.2,...,1.0},
        ytick={0,20,...,100},
        ylabel shift={-5pt},        
        grid = major,
        major grid style={dotted,gray},
    ]

\addplot[color=red, solid, mark options={solid, fill=red, draw=black}, mark=triangle*, line width=1pt, mark size=3pt, visualization depends on=\thisrow{alignment} \as \alignment, nodes near coords, point meta=explicit symbolic,
    every node near coord/.style={anchor=\alignment, font=\small
    }] 
table [meta index=2]  {
x   y       label   alignment
0	12.34	12.34		-60
0.1	15.25	{}	-20
0.3	23.83	23.83	-20
0.5	36.32	36.32	-20
0.7 56.19	56.19		-20
0.9 86.39	86.39		-20
1.0	106.56	106.56		0
	};
    \addlegendentry{\knnlm{}}

\addplot[color=blue, mark options={solid, fill=blue, draw=black}, mark=square*, line width=1pt, mark size=3pt, visualization depends on=\thisrow{alignment} \as \alignment, nodes near coords, point meta=explicit symbolic,
    every node near coord/.style={anchor=\alignment, 
   		font=\small
    }] 
table [meta index=2]  {
x   y       label   alignment
0	12.01	12.01	5
0.1	12.67	{}		-90
0.3	16.22	16.22		-20
0.5	24.50	24.50		-180
0.7 40.70	40.70		140
0.9 74.38	74.38		-20
1.0	106.56	{}		0
	};
    \addlegendentry{\adaptret{}}

\addplot[color=ao, mark options={solid, fill=ao, draw=black}, mark=*, line width=1pt, mark size=3pt, visualization depends on=\thisrow{alignment} \as \alignment, nodes near coords, point meta=explicit symbolic,
    every node near coord/.style={anchor=\alignment, font=\small
    }] 
table [meta index=2]  {
x   y       label   alignment
0.00	10.49	\textbf{10.49}	90
0.35	10.70	10.70	90
0.45	10.81	{}	0
0.53	10.89	10.89	90
0.61	10.97	{}	0
0.73	11.25	11.25	-60
0.83	11.98	11.98	-110
1.0	106.56	{}		0
	};
    \addlegendentry{\ourmodel{} (this work)}

	\end{axis}
\end{tikzpicture}
\caption{Experiments for domain adaptation, where the datastore is constructed from Law-MT.}
\label{fig:law}
\end{figure}
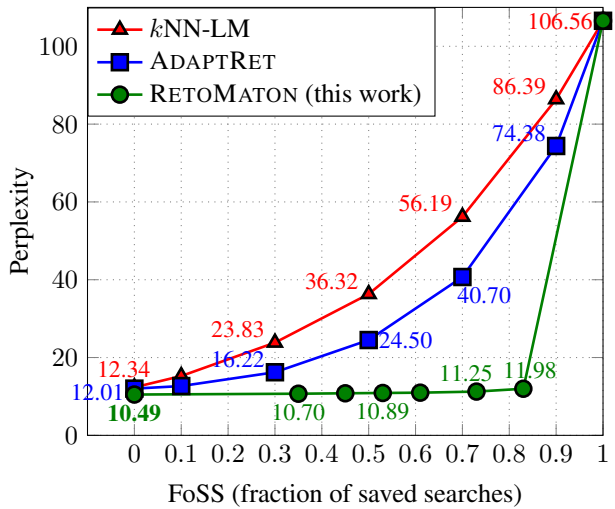

%% file: fig_fintuned.tex
\begin{table}[h!]
	\caption{Experiments using a base LM that was fine-tuned on Law-MT. 
	Numbers denote perplexity, and the relative reduction over the fine-tuned LM is shown in parentheses.
	}
	\label{fig:law-finetuned}
	\centering
	\begin{tabular}{lrrrrr}
	  \toprule
	  	Model  & \multicolumn{2}{c}{FoSS$=$0} & \multicolumn{2}{c}{FoSS$=$0.5} \\
	  \midrule
fine-tuned LM		&  \multicolumn{4}{c}{8.61} \\
\knnlm{} &  7.93 & ($\downarrow$7.9\%) & 8.25  & ($\downarrow$4.2\%) \\
\adaptret{} & 7.81  &  ($\downarrow$9.2\%) & 7.91  & ($\downarrow$8.1\%)\\
\midrule
\ourmodel{}  & \textbf{7.10} & \textbf{($\downarrow$17.5\%)} & \textbf{7.15} & \textbf{($\downarrow$17.0\%)} \\
	  \bottomrule
	\end{tabular}
	\end{table}

%% file: 06_ablation.tex
\input{kmeans_fig.tex}
\input{07_examples.tex}

\input{07_long_example.tex}
\section{Ablation Study}
\label{sec:analysis}
\paragraph{Pointers vs. clustering}
The main two contributions in \ourmodel{} are the use of pointers and the clustering.
Here, we tease apart the contribution of each of these.

The \emph{w/o clustering} model is an ablation of \ourmodel{}, which spares the clustering preprocessing step, and uses only pointers. 
 As shown in \Cref{fig:wiki}, even the \emph{w/o clustering} model achieves lower perplexity than the other baselines, with a lowest perplexity of 16.12 at FoSS$=$0.
Up to FoSS$=$0.4, \emph{w/o clustering} performs only slightly worse than the base \ourmodel{}.
Starting from FoSS$=$0.7, the \emph{w/o clustering} model almost consolidates with \adaptret{}.

From these experiments, we conjecture that \ourmodel{}'s performance in few saved searches (FoSS$<$0.4) stems mostly from keeping pointers, which provides the most significant boost. 
Starting from FoSS$=$0.7, the gap between \emph{w/o clustering} and the base \ourmodel{} shows the contribution of clustering, 
which allows longer sequences of consecutive steps without performing \knn{} search.

\paragraph{Clustering Granularity} The only hyperparameter that \ourmodel{} introduces is the choice of the number of clusters.
A higher number of clusters 
results in smaller, fine-grained clusters. %
A low number of clusters results in larger, coarse-grained clusters, where every cluster is possibly more noisy.

Here, we vary the number of clusters and also experiment with the ``greedy'' clustering algorithms from \citet{he2021efficient}. The advantage of the greedy algorithm is that it is computationally cheaper: it requires searching for each datastore entry's nearest neighbors, and then performing a single pass of merging, while $k$-means requires multiple iterations over the entire datastore. 

The results of these experiments are shown in \Cref{fig:analysis_wiki} for \wiki{} and \Cref{fig:analysis_law} for Law-MT. 
As shown in \Cref{fig:analysis_wiki}, $k$$=$500K means and $k$$=$1M means achieve similar perplexities, while $k$$=$100K  is too coarse-grained. The \emph{greedy} algorithm presents an interesting tradeoff, as it achieves lower perplexity than the others at FoSS$=$0, but degrades as FoSS increases, since each cluster is smaller. 
\Cref{fig:analysis_law} shows a similar tradeoff in Law-MT: the most fine-grained clustering using $k$$=$400K means performs best for FoSS$=$0, but achieves a higher perplexity than others at FoSS$>$0.7. Additional clustering runs are shown in \Cref{app:analysis}.

%% file: kmeans_fig.tex
\definecolor{mypurple}{HTML}{AB30C4}

\begin{figure*}
\begin{minipage}[t][][b]{0.48\textwidth}	
		\begin{tikzpicture}[scale=1]
	\begin{axis}[
		xlabel={FoSS (fraction of saved searches)},
		ylabel={Perplexity},
		ylabel near ticks,
        legend style={at={(0,1)},anchor=north west,mark size=2pt,
        	}, %
        legend cell align={left},
        xmin=0.0, xmax=0.9,
        ymin=16.09, ymax=17,
        xtick={0.1,0.2,...,1.0},
        ytick={16.1,16.2,...,17.0},
        ylabel shift={-5pt},        
        grid = major,
        major grid style={dotted,gray},
    ]

\addplot[color=gray, solid, mark options={solid, fill=white, draw=black}, mark=otimes, line width=1pt, mark size=3pt, visualization depends on=\thisrow{alignment} \as \alignment, nodes near coords, point meta=explicit symbolic,
    every node near coord/.style={anchor=\alignment, font=\footnotesize}] 
table [meta index=2]  {
x   y       label   alignment
0	16.13	{}		0
0.20	16.20	{}		0
0.24	16.23	{}		0
0.29	16.26	{}		0
0.37	16.33	{}		0
0.43	16.39	{}		0	
0.50	16.48	{}		0
0.55	16.55	{}		0
0.59	16.63	{}		0
0.68	16.91	{}	0
	};
    \addlegendentry{w/o clustering}

\addplot[color=red, solid, mark options={solid, fill=red, draw=black}, mark=triangle*, line width=1pt, mark size=2pt, visualization depends on=\thisrow{alignment} \as \alignment, nodes near coords, point meta=explicit symbolic,
    every node near coord/.style={anchor=\alignment, font=\footnotesize}] 
table [meta index=2]  {
x   y       label   alignment
0	16.11	{}		0
0.29	16.21	{}	0
0.34	16.23	{}	0
0.39	16.26	{}	0
0.48	16.30	{}		0
0.55	16.33	{}	0
0.64	16.38	{}	0
0.70	16.43	{}	0
0.75	16.48	{}	0
0.83	16.69	{}	0
	};
    \addlegendentry{$k$$=$1M means}

\addplot[color=blue, mark options={solid, fill=blue, draw=black}, mark=square*, line width=1pt, mark size=2pt, visualization depends on=\thisrow{alignment} \as \alignment, nodes near coords, point meta=explicit symbolic,
    every node near coord/.style={anchor=\alignment, font=\footnotesize}] 
table [meta index=2]  {
x   y       label   alignment
0.0	16.11	{}	0
0.30	16.23	{}	0
0.35	16.25	{}	0
0.40	16.28	{}	0
0.48	16.32	{}	0
0.56	16.35	{}	0
0.65	16.41	{}	0
0.70	16.46	{}	0
0.75	16.51	{}	0
0.84	16.70	{}	0
	};
    \addlegendentry{$k$$=$500K means}

\addplot[color=ao, mark options={solid, fill=ao, draw=black}, mark=*, line width=1pt, mark size=2pt, visualization depends on=\thisrow{alignment} \as \alignment, nodes near coords, point meta=explicit symbolic,
    every node near coord/.style={anchor=\alignment, font=\small
    }] 
table [meta index=2]  {
x   y       label   alignment
0	16.1	{}	0
0.31	16.26	{}	0
0.41	16.32	{}	0
0.49	16.37	{}	0
0.56	16.42	{}	0
0.64	16.49	{}	0
0.70	16.56	{}	0
0.75	16.63	{}	0
0.84	16.82	{}	0
	};
    \addlegendentry{$k$$=$100K means}

\addplot[color=orange, mark options={solid, fill=orange, draw=black}, mark=diamond*, line width=1pt, mark size=2pt, visualization depends on=\thisrow{alignment} \as \alignment, nodes near coords, point meta=explicit symbolic,
    every node near coord/.style={anchor=\alignment, font=\small
    }] 
table [meta index=2]  {
x   y       label   alignment
0	16.09	{}	0
0.26	16.18	{}	0
0.30	16.20	{}	0
0.36	16.23	{}	0
0.45	16.28	{}	0
0.52	16.33	{}	0
0.59	16.40	{}	0
0.65	16.48	{}	0
0.69	16.57	{}	0
0.75	16.83	{}	0
	};
    \addlegendentry{greedy, 21M}

	\end{axis}
\end{tikzpicture}
\caption{Analysis of the number of clusters on the validation set of \wiki{}. Additional clustering runs can be found in \Cref{app:analysis} (\Cref{fig:app:analysis_wiki}).}
\label{fig:analysis_wiki}
\end{minipage}
\hfill
\begin{minipage}[t][][b]{0.48\textwidth}	
	\begin{tikzpicture}[scale=1]
	\begin{axis}[
		xlabel={FoSS (fraction of saved searches)},
		ylabel={Perplexity},
		ylabel near ticks,
        legend style={at={(0,1)},anchor=north west,mark size=2pt,
        	}, %
        legend cell align={left},
        xmin=0, xmax=0.9,
        ymin=10.6, ymax=12,
        xtick={0.0, 0.1,0.2,...,1.0},
        ytick={10.0, 10.2,...,16},
        ylabel shift={-5pt},        
        grid = major,
        major grid style={dotted,gray},
    ]

\addplot[color=gray, solid, mark options={solid, fill=white, draw=black}, mark=otimes, line width=1pt, mark size=3pt, visualization depends on=\thisrow{alignment} \as \alignment, nodes near coords, point meta=explicit symbolic,
    every node near coord/.style={anchor=\alignment, font=\footnotesize}] 
table [meta index=2]  {
x   y       label   alignment
0	10.85	{}		0
0.29	11.02	{}		0
0.38	11.20	{}		0
0.45	11.46	{}		0
0.51	11.79	{}		0
0.59	12.29	{}		0
0.64	12.69	{}		0
0.69	13.72	{}		0
0.76	16.83	{}		0
	};
    \addlegendentry{w/o clustering}

\addplot[color=red, solid, mark options={solid, fill=red, draw=black}, mark=triangle*, line width=1pt, mark size=2pt, visualization depends on=\thisrow{alignment} \as \alignment, nodes near coords, point meta=explicit symbolic,
    every node near coord/.style={anchor=\alignment, font=\footnotesize}] 
table [meta index=2]  {
x   y       label   alignment
0	10.64	{}		0
0.40	10.80	{}		0
0.49	10.87	{}		0
0.57	10.96	{}		0
0.64	11.03	{}		0
0.71	11.14	{}		0
0.75	11.26	{}		0
0.80	11.41	{}		0
0.84	11.80	{}		0
	};
    \addlegendentry{$k$$=$100K means}

\addplot[color=blue, mark options={solid, fill=blue, draw=black}, mark=square*, line width=1pt, mark size=2pt, visualization depends on=\thisrow{alignment} \as \alignment, nodes near coords, point meta=explicit symbolic,
    every node near coord/.style={anchor=\alignment, font=\footnotesize}] 
table [meta index=2]  {
x   y       label   alignment
0.0	10.61	{}		0
0.39	10.75	{}		0
0.48	10.81	{}		0
0.57	10.88	{}		0
0.64	10.96	{}		0
0.71	11.09	{}		0
0.75	11.21	{}		0
0.80	11.33	{}		0
0.84	11.79	{}		0
	};
    \addlegendentry{$k$$=$200K means}

\addplot[color=ao, mark options={solid, fill=ao, draw=black}, mark=*, line width=1pt, mark size=2pt, visualization depends on=\thisrow{alignment} \as \alignment, nodes near coords, point meta=explicit symbolic,
    every node near coord/.style={anchor=\alignment, font=\small
    }] 
table [meta index=2]  {
x   y       label   alignment
0	10.60	{}		0
0.38	10.72	{}		0
0.48	10.76	{}		0
0.56	10.83	{}		0
0.63	10.91	{}		0
0.70	11.06	{}		0
0.74	11.21	{}		0
0.79	11.39	{}		0
0.84	11.99	{}		0
	};
    \addlegendentry{$k$$=$400K means}

	\end{axis}
\end{tikzpicture}
\caption{Analysis of the number of clusters on the validation set of Law-MT. A larger version of this figure can be found in \Cref{app:analysis} (\Cref{fig:app:analysis_law}).}
\label{fig:analysis_law}
\end{minipage}
\end{figure*}

%% file: 07_examples.tex
\begin{table*}[t]
\centering
\small
\begin{tabular}{lll}
\toprule
length=3 & length=6 & length=10 \\
\midrule 
\emph{, and the} & \emph{the Streets Have No Name "}  & \emph{. As a result , there was not a single} \\
\emph{, but the} & \emph{Department of Transportation ( MDOT )"} & \emph{and some were moved to new locations . Before its}   \\
\emph{roughly bounded by} & \emph{In the United States , }  & \emph{but it was not until the following day that a}  \\
\emph{in the first} & \emph{the end of the song ,}  & \emph{the end of the Second World War was completed in}  \\
\emph{a number of} & \emph{not occur until 27 May 1915}  & \emph{to lack of evidence ; however , the decision was}  \\
\emph{, when the} & \emph{fired the shots that caused the}   & \emph{the Streets Have No Name ' is more like the} \\
\bottomrule
\end{tabular}	
\caption{Some of the sequences from the \wiki{} validation set that our automaton captured without performing \knn{} search. We selected length=10 sequences that did not appear in the training data.}
\label{fig:examples}
\end{table*}

%% file: 07_long_example.tex
\begin{figure*}[t]
\centering
\small
\begin{tabular}{>{\arraybackslash}m{16.5cm}}
        \toprule %
 {Training Sequence} from: \url{https://en.wikipedia.org/wiki/Oh,_What_a_Knight!} \\ \midrule %
 \emph{\textbf{The writer of the scenario is unknown , but it was most likely Lloyd Lonergan . He was an experienced newspaperman employed by The New York Evening World while ...
A surviving film still gives the possibility of identifying} three of the actors in the film ... } \\
\midrule
 {Validation Sequence} from: \url{https://en.wikipedia.org/wiki/Home_Made_Mince_Pie} \\ 
                     \midrule 
 \emph{\textbf{The writer of the scenario is unknown , but it was most likely Lloyd Lonergan . He was an experienced newspaperman employed by The New York Evening World while ...
A surviving film still gives the possibility of identifying} eight actors ...} \\
\bottomrule %
\end{tabular}
    \caption{An example for a 236-token long sequence that \ourmodel{} was able to capture from the \wiki{} training set and apply to the validation set. 
    Although most of the paragraph appears as-is in both  sets, they have different endings, as these are articles about different silent films from 1910.  This shows how \ourmodel{} allows to dynamically retrieve chunks of text from the training data using a single \knn{} search, rather than a single token at a time as \knnlm{}.}
    \label{fig:long-example}
\end{figure*}

%% file: 07_qualitative.tex
\label{sec:analysis}
\section{Qualitative Analysis}
What are the sequences of tokens that \ourmodel{} captured and predicted consecutively, without \knn{} search?

\Cref{fig:examples} shows examples of sequences among those that were given the highest probability ($p_{\text{auto}}$ in \Cref{eq:pauto}) from the validation set of \wiki{}.
Naturally, short sequences (length=3) are usually common 3-grams such as ``\emph{, and the}''. As length increases (to 6 and 10 tokens), the sequences become more specific; for example, two of them contain part of the name of the song ``\emph{Where the Streets Have No Name}'' by the band U2. 
Nevertheless, we selected sequences into the list of length=10 such that \emph{none of them appeared as-is in the training set}, to show that \ourmodel{} does not only memorize n-grams, but instead, clustering allows it to compose multiple small n-grams into longer n-grams.

\Cref{fig:long-example} shows a 217-token long passage that appears in both training and validation sets, but with different endings.  \ourmodel{} can predict this passage consecutively, without performing search. This shows how \ourmodel{} can retrieve single tokens from the datastore, but it also can adaptively construct much longer chunks, if needed.

\Cref{fig:histo_lengths} (\Cref{app:analysis}) shows a histogram of the lengths of these sequences. 
The vast majority (98\%) of the validation set tokens are included in n-grams having n$>$1, which either started or continued an automaton traversal.

\Cref{fig:clusters} shows a random sample of states and transitions from the automaton construced from the training set of \wiki{}. Further exploration of the automaton can be performed using our publicly available code. %

%% file: 08_related.tex
\section{Related Work}
\label{sec:related}
\input{cluster_fig.tex}

\subsection{Comparison to \badaptret{} \cite{he2021efficient}}
\label{subsec:comparehe}
The closest work to ours is \adaptret{} \cite{he2021efficient}. \adaptret{} saves \knn{} searches as well, but it suffers from conceptual weaknesses compared to our work:

\textbf{When the model performs $\bm{k}$NN search}:
\adaptret{} uses only the neighbors retrieved by the \knn{} search.
\ourmodel{} uses the set of retrieved nearest neighbors as well, but also includes the remaining pointers from the previous time step (\Cref{eq:restart}). Apparently, these remaining pointers have a higher likelihood of predicting the correct token than the general set of retrieved nearest neighbors.

\textbf{When the model does not perform $\bm{k}$NN search}:
\adaptret{} skips the interpolation of $p_{k\text{NN}}$ with $p_{LM}$, and uses $p_{LM}$ solely.
In contrast, \ourmodel{} %
 still computes the interpolation of $p_{\text{auto}}$ with $p_{LM}$, thanks to its pointers.
As an interesting direction for future work, we expect that 
learning a dynamic interpolation factor $\lambda$, similarly to \adaptret{}, will even further improve \ourmodel{}'s results.

\paragraph{Data Efficiency} \adaptret{} requires training 
its MLP  
on a dataset that is disjoint from the corpus that the datastore was built from, to prevent 
overfitting. Thus, \citeauthor{he2021efficient} had to train their MLP \emph{on the validation set}, which is not data-efficient -- spending 90\% of the original validation set for additional training. In contrast, our approach is completely \emph{unsupervised}, and thus does not require additional data.

\subsection{Retrieval and Neuro-Symbolic Methods}
\paragraph{Granularity of Retrieval} 
While \citet{khandelwal2020generalization} and \citet{yogatama2021adaptive} retrieve a token at a time step, other work retrieved a sentence \cite{hashimoto2018retrieve,gu2018search,rubin2021learning}, a prototype \cite{guu2018generating,he2020learning}, or a chunk \cite{guu2020realm,borgeaud2021improving}.
\ourmodel{} implicitly generalizes these approaches by dynamically constructing  the retrieved sequence, essentially being able to retrieve individual tokens as well as constructing search-free longer passages.

\paragraph{Hybrid Models}
Combining n-grams \cite{neubig2016generalizing} and automata \cite{rijhwani2021lexically} with neural language models has
usually led to ``static'', count-based, transitions weights.
In contrast, states in our automaton are based on hidden representations of the  neural LM, which allows \ourmodel{} to
\emph{dynamically} weigh transitions. 
Other work scored automata with RNNs
\cite{rastogi2016weighting,lin2019neural}, or constructed RNNs from automata \cite{schwartz2018bridging,peng2018rational}; 
\ourmodel{} differs from these approaches by providing with \emph{retrieved instances} from a datastore, instead of enforcing structure on the neural LM.

\paragraph{Leveraging Structure in \knnlm{}s} 
\citet{meng2022gnnlm} train a graph neural network (GNN) on top of the encoded test contexts and the retrieved examples. %
Differently from our work, their GNN requires additional training, while our approach is \emph{completely unsupervised}. 
Applying \ourmodel{} on top of their approach 
is an interesting future direction 
which is expected to further improve their results.

\paragraph{Automata Extraction} 
The extraction of automata from neural networks  goes back to \citet{giles1992learning} and \citet{omlin1996extraction}, mainly for synthetic and simple regular languages.
Later, \citet{weiss2018extracting,weiss2019learning} scaled up the extraction to larger GRU and LSTM architectures. In this work, we do \emph{not only extract} an automaton, but also combine it with a neural LM to improve the LM's accuracy.

%% file: cluster_fig.tex
\newcommand{\noendgray}{gray!20!white}
\newcommand{\noend}[1]{
    \draw [-{Latex[length=1mm, line width=1pt]},out=0,in=180,draw=\noendgray] (#1.east) -- ++(4mm,0);
    \draw [-{Latex[length=1mm, line width=1pt]},out=0,in=180,draw=\noendgray] (#1.east) -- ++(4mm,4mm);
    \draw [-{Latex[length=1mm, line width=1pt]},out=0,in=180,draw=\noendgray] (#1.east) -- ++(4mm,-4mm);
}

\begin{figure*}[t]
    \begin{tikzpicture}[node distance=2cm]
       \centering
       \footnotesize 
        \node [circle,draw=myred,inner sep=2pt,fill=Apricot!20!white,line width=1pt] (node10) {};
        \node [circle,draw=myred, right of=node10, inner sep=2pt,fill=Apricot!20!white,line width=1pt] (node17308) {};
        \node [circle,draw=myred,below of=node17308, inner sep=2pt,fill=Apricot!20!white,line width=1pt] (node78017) {};
        \node [circle,draw=myred,below of=node78017, inner sep=2pt,fill=Apricot!20!white,line width=1pt] (node384715) {};
        \node [circle,draw=myred,below of=node384715, inner sep=2pt,fill=Apricot!20!white,line width=1pt] (node977853) {};
        \node [circle,draw=myred,below of=node977853, inner sep=2pt,fill=Apricot!20!white,line width=1pt] (node997169) {};
        \node [circle,draw=myred,below of=node997169, inner sep=2pt,fill=Apricot!20!white,line width=1pt] (node840109) {};

        \node [circle,draw=myred, right of=node17308, inner sep=2pt,fill=Apricot!20!white,line width=1pt] (node9406) {};
        \node [circle,draw=myred, below of=node9406, inner sep=2pt,fill=Apricot!20!white,line width=1pt] (node14238) {};
        \node [circle,draw=myred, below of=node14238, inner sep=2pt,fill=Apricot!20!white,line width=1pt] (node586385) {};
        \node [circle,draw=myred, below of=node586385, inner sep=2pt,fill=Apricot!20!white,line width=1pt] (node795349) {};
        \node [circle,draw=myred, below of=node795349, inner sep=2pt,fill=Apricot!20!white,line width=1pt] (node64500) {};
        \node [circle,draw=myred, below of=node64500, inner sep=2pt,fill=Apricot!20!white,line width=1pt] (node129929) {};
        
        \node [circle,draw=myred, right of=node9406, inner sep=2pt,fill=Apricot!20!white,line width=1pt] (node123447) {};
        \node [circle,draw=myred, below of=node123447, inner sep=2pt,fill=Apricot!20!white,line width=1pt] (node158276) {};
        \node [circle,draw=myred, below of=node158276, inner sep=2pt,fill=Apricot!20!white,line width=1pt] (node643439) {};
        \node [circle,draw=myred, below of=node643439, inner sep=2pt,fill=Apricot!20!white,line width=1pt] (node337565) {};
        \node [circle,draw=myred, below of=node337565, inner sep=2pt,fill=Apricot!20!white,line width=1pt] (node464381) {};
        \node [circle,draw=myred, below of=node464381, inner sep=2pt,fill=Apricot!20!white,line width=1pt] (node876672) {};

        \node [circle,draw=myred, right of=node123447, inner sep=2pt,fill=Apricot!20!white,line width=1pt] (node43962) {}; %
        \node [circle,draw=myred, below of=node43962, inner sep=2pt,fill=Apricot!20!white,line width=1pt] (node61735) {}; %
        \node [circle,draw=myred, below of=node61735, inner sep=2pt,fill=Apricot!20!white,line width=1pt] (node490105) {}; %
        \node [circle,draw=myred, below of=node490105, inner sep=2pt,fill=Apricot!20!white,line width=1pt] (node8459) {}; %
        \node [circle,draw=myred, below of=node8459, inner sep=2pt,fill=Apricot!20!white,line width=1pt] (node210791) {}; %
        \node [circle,draw=myred, below of=node210791, inner sep=2pt,fill=Apricot!20!white,line width=1pt] (node633526) {}; %
        
        \node [circle,draw=myred, right of=node43962, inner sep=2pt,fill=Apricot!20!white,line width=1pt] (node132484) {}; %
        \node [circle,draw=myred, below of=node132484, inner sep=2pt,fill=Apricot!20!white,line width=1pt] (node46618) {}; %
        \node [circle,draw=myred, below of=node46618, inner sep=2pt,fill=Apricot!20!white,line width=1pt] (node618143) {}; %
        \node [circle,draw=myred, below of=node618143, inner sep=2pt,fill=Apricot!20!white,line width=1pt] (node377433) {}; %
        \node [circle,draw=myred, below of=node377433, inner sep=2pt,fill=Apricot!20!white,line width=1pt] (node667550) {}; %
        \node [circle,draw=myred, below of=node667550, inner sep=2pt,fill=Apricot!20!white,line width=1pt] (node569167) {}; %

        \node [circle,draw=myred, right of=node132484, inner sep=2pt,fill=Apricot!20!white,line width=1pt] (node165941) {}; %
        \node [circle,draw=myred, below of=node165941, inner sep=2pt,fill=Apricot!20!white,line width=1pt] (node320511) {}; %
        \node [circle,draw=myred, below of=node320511, inner sep=2pt,fill=Apricot!20!white,line width=1pt] (node709735) {}; %
        \node [circle,draw=myred, below of=node709735, inner sep=2pt,fill=Apricot!20!white,line width=1pt] (node12903) {}; %
        \node [circle,draw=myred, below of=node12903, inner sep=2pt,fill=Apricot!20!white,line width=1pt] (node427831) {}; %
        \node [circle,draw=myred, below of=node427831, inner sep=2pt,fill=Apricot!20!white,line width=1pt] (node769008) {}; %
        
        \node [circle,draw=myred, right of=node165941, inner sep=2pt,fill=Apricot!20!white,line width=1pt] (node121469) {}; %
        \node [circle,draw=myred, below of=node121469, inner sep=2pt,fill=Apricot!20!white,line width=1pt] (node26606) {}; %
        \node [circle,draw=myred, below of=node26606, inner sep=2pt,fill=Apricot!20!white,line width=1pt] (node879766) {}; %
        \node [circle,draw=myred, below of=node879766, inner sep=2pt,fill=Apricot!20!white,line width=1pt] (node697325) {}; %
        \node [circle,draw=myred, right of=node769008, inner sep=2pt,fill=Apricot!20!white,line width=1pt] (node674949) {}; %

        \draw [-{Latex[length=2mm, line width=1pt]},out=0,in=180] (node10.east) to node[pos=0.5,right,fill=white] {his} (node17308.west) ;
        \draw [-{Latex[length=2mm, line width=1pt]},out=0,in=180] (node10.east) to node[right] {their} (node78017.west) ;
        \draw [-{Latex[length=2mm, line width=1pt]},out=-30,in=140] (node10.east) to node[pos=0.7,fill=white] {her/his} (node384715.west) ;
        \draw [-{Latex[length=2mm, line width=1pt]},out=-60,in=150] (node10.east) to node[pos=0.6,fill=white] {Cooper} (node977853.west) ;
        \draw [-{Latex[length=2mm, line width=1pt]},out=-80,in=150] (node10.east) to node[fill=white] {what} (node997169.west) ;
        \draw [-{Latex[length=2mm, line width=1pt]},out=-90,in=150] (node10.south) to node[pos=0.7,fill=white] {salvation} (node840109.west) ;

        \draw [-{Latex[length=2mm, line width=1pt]},out=0,in=180] (node17308.east) to node[pos=0.2,right,fill=white] {strange} (node9406.west) ;
        \draw [-{Latex[length=2mm, line width=1pt]},out=0,in=180] (node17308.east) to node[right] {death} (node14238.west) ;
        \draw [-{Latex[length=2mm, line width=1pt]},out=-30,in=140] (node17308.east) to node[pos=0.7,fill=white] {past} (node586385.west) ;
        \draw [-{Latex[length=2mm, line width=1pt]},out=0,in=180] (node977853.east) to node[pos=0.2,right,fill=white] {Draper} (node795349.west) ;
        \draw [-{Latex[length=2mm, line width=1pt]},out=0,in=180] (node977853.east) to node[right] {Clarke} (node64500.west) ;
        \draw [-{Latex[length=2mm, line width=1pt]},out=-30,in=140] (node977853.east) to node[pos=0.7,fill=white] {'s} (node129929.west) ;
        
        \draw [-{Latex[length=2mm, line width=1pt]},out=0,in=180] (node9406.east) to node[pos=0.2,right,fill=white] {dreams} (node123447.west) ;
        \draw [-{Latex[length=2mm, line width=1pt]},out=0,in=180] (node9406.east) to node[right] {phenomena} (node158276.west) ;
        \draw [-{Latex[length=2mm, line width=1pt]},out=-30,in=140] (node9406.east) to node[pos=0.7,fill=white] {creatures} (node643439.west) ;
        \draw [-{Latex[length=2mm, line width=1pt]},out=30,in=-140] (node129929.east) to node[pos=0.7,fill=white] {musical} (node337565.west) ;
        \draw [-{Latex[length=2mm, line width=1pt]},out=0,in=180] (node129929.east) to node[right] {direction} (node464381.west) ;
        \draw [-{Latex[length=2mm, line width=1pt]},out=0,in=180] (node129929.east) to node[below] {performance} (node876672.west) ;

        \draw [-{Latex[length=2mm, line width=1pt]},out=30,in=-140] (node643439.east) to node[pos=0.8,fill=white] {from} (node43962.west) ;
        \draw [-{Latex[length=2mm, line width=1pt]},out=0,in=180] (node643439.east) to node[right] {such} (node61735.west) ;
        \draw [-{Latex[length=2mm, line width=1pt]},out=0,in=180] (node643439.east) to node[below] {called} (node490105.west) ;
        \draw [-{Latex[length=2mm, line width=1pt]},out=0,in=180] (node337565.east) to node[above] {soundtrack} (node8459.west) ;
        \draw [-{Latex[length=2mm, line width=1pt]},out=0,in=180] (node337565.east) to node[right] {styles} (node210791.west) ;
        \draw [-{Latex[length=2mm, line width=1pt]},out=-30,in=140] (node337565.east) to node[pos=0.8,fill=white] {score} (node633526.west) ;

        \draw [-{Latex[length=2mm, line width=1pt]},out=30,in=-140] (node490105.east) to node[pos=0.8,fill=white] {the} (node132484.west) ;
        \draw [-{Latex[length=2mm, line width=1pt]},out=0,in=180] (node490105.east) to node[right] {} (node46618.west) ;
        \draw [-{Latex[length=2mm, line width=1pt]},out=0,in=180] (node490105.east) to node[below] {"} (node618143.west) ;
        \draw [-{Latex[length=2mm, line width=1pt]},out=30,in=-140] (node633526.east) to node[pos=0.7,fill=white] {was} (node377433.west) ;
        \draw [-{Latex[length=2mm, line width=1pt]},out=0,in=180] (node633526.east) to node[right] {for} (node667550.west) ;
        \draw [-{Latex[length=2mm, line width=1pt]},out=0,in=180] (node633526.east) to node[below] {contains} (node569167.west) ;
        \draw [-{Latex[length=2mm, line width=1pt]},out=90,in=180] (node618143.north) to node[below,fill=white] {Pokémon} (node46618.south) ;

        \draw [-{Latex[length=2mm, line width=1pt]},out=30,in=-140] (node618143.east) to node[pos=0.8,fill=white] {hairballs} (node165941.west) ;
        \draw [-{Latex[length=2mm, line width=1pt]},out=0,in=180] (node618143.east) to node[right] {titans} (node320511.west) ;
        \draw [-{Latex[length=2mm, line width=1pt]},out=0,in=180] (node618143.east) to node[below] {monster} (node709735.west) ;
        \draw [-{Latex[length=2mm, line width=1pt]},out=0,in=180] (node377433.east) to node[above] {composed} (node12903.west) ;
        \draw [-{Latex[length=2mm, line width=1pt]},out=0,in=180] (node377433.east) to node[right] {written} (node427831.west) ;
        \draw [-{Latex[length=2mm, line width=1pt]},out=-30,in=140] (node377433.east) to node[pos=0.8,fill=white] {produced} (node769008.west) ;

        \draw [-{Latex[length=2mm, line width=1pt]},out=0,in=180] (node165941.east) to node[fill=white] {"} (node121469.west) ;
        \draw [-{Latex[length=2mm, line width=1pt]},out=0,in=180] (node320511.east) to node[fill=white] {"} (node121469.west) ;
        \draw [-{Latex[length=2mm, line width=1pt]},out=0,in=180] (node709735.east) to node[fill=white] {puppets} (node26606.west) ;
        \draw [-{Latex[length=2mm, line width=1pt]},out=0,in=180] (node709735.east) to node[below] {sharks} (node879766.west) ;
        
        \draw [-{Latex[length=2mm, line width=1pt]},out=0,in=180] (node12903.east) to node[pos=0.5,fill=white] {by} (node697325.west) ;
        \draw [-{Latex[length=2mm, line width=1pt]},out=0,in=180] (node427831.east) to node[left,fill=white] {by} (node697325.west) ;
        \draw [-{Latex[length=2mm, line width=1pt]},out=0,in=-140] (node769008.east) to node[fill=white] {by} (node697325.west) ;
        
        \draw [-{Latex[length=2mm, line width=1pt]},out=0,in=180] (node769008.east) to node[pos=0.5,fill=white] {during} (node674949.west) ;

        \noend{node78017}
        \noend{node384715}
        \noend{node997169}
        \noend{node840109}
        
        \noend{node14238}
        \noend{node586385}
        \noend{node795349}
        \noend{node64500}

        \noend{node123447}
        \noend{node158276}
        \noend{node464381}
        \noend{node876672}
        \noend{node43962}
        \noend{node61735}
        \noend{node210791}
        \noend{node132484}
        \noend{node8459}
        \noend{node46618}
        \noend{node667550}
        \noend{node569167}
        \noend{node121469}
        \noend{node26606}
        \noend{node879766}
        \noend{node697325}
        \noend{node674949}

    \end{tikzpicture}
    \caption{A random sample of the automaton constructed from the training set of \wiki{}}
    \label{fig:clusters}
\end{figure*}
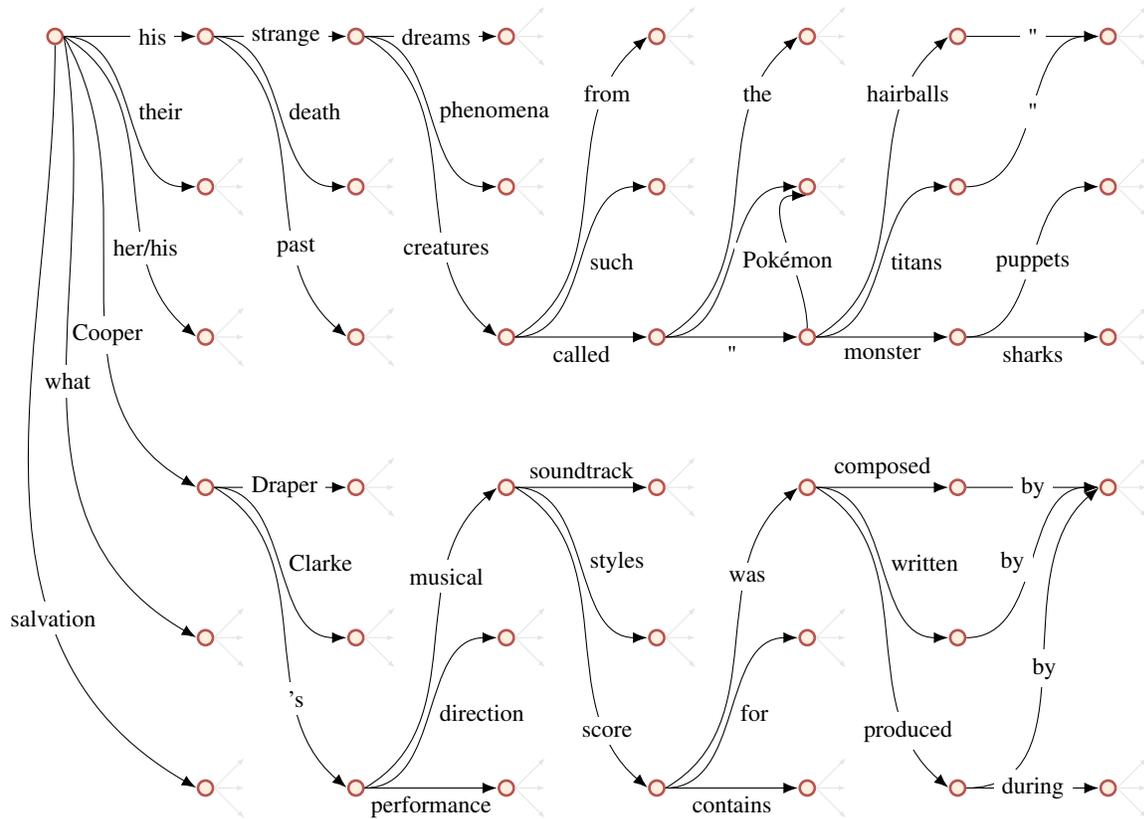

%% file: 09_conclusion.tex
\section{Conclusion}
\label{sec:conclusion}
We presented \ourmodel{} -- retrieval automaton.
\ourmodel{} approximates a \knn{} search over an external corpus by clustering similar neighbors into automaton states, and keeping pointers from previously found neighbors, which form the transition between  states.
This results in a weighted finite automaton, which allows approximating the nearest neighbors in most of the time steps, instead of performing a \knn{} search at every step.

Empirically, traversing the automaton at inference time saves up to 83\% of the \knn{} searches in both in-domain and domain adaptation settings, and reduces the perplexity of strong LMs, even after they were fine-tuned.

These results suggest a promising direction for the neuro-symbolic synergy of neural models with symbolic automata.
We believe that the principles and the methods presented in this paper are also applicable to other R-LMs, including phrase- and chunk-based retrieval models.
To these ends, 
we make all our code, data, and models publicly available. %

%% file: 10_acks.tex
\section*{Acknowledgments}
We thank Lucio Dery and Vincent Hellendoorn for the helpful discussions and thorough feedback. We are also grateful to the anonymous reviewers for their useful comments and suggestions.

%% file: appendix_main.tex
\appendix
\onecolumn

\section{Intuition for $\bm{\tau}$}
\label{app:tau}
For brevity, we repeat \Cref{eq:restart} here:
\begin{align*}
	&\mathcal{S}^{\left(t+1\right)}=  
	 \begin{cases}
		\hat{\delta}\left(\mathcal{S}^{\left(t\right)}, w^{\left(t\right)}\right) &  \abs{\hat{\delta}\left(\mathcal{S}^{\left(t\right)}, w^{\left(t\right)}\right)}\geq\tau \\
		\hat{\delta}\left(\mathcal{S}^{\left(t\right)}, w^{\left(t\right)}\right) \cup \bigcup\limits_{e\in \mathcal{N}^{\left(t+1\right)}}\pi\left(e\right)
		& \mathrm{otherwise}
	\end{cases} \nonumber
\end{align*} 

Intuitively, the larger $\abs{\hat{\delta}\left(\mathcal{S}^{\left(t\right)}, w^{\left(t\right)}\right)}$ is, the more ``correct entries'' -- entries whose values were equal to $w^{\left(t\right)}$ that we had at time $t$, and the more we can rely on their pointers at time $t+1$. 
Thus, if $\abs{\hat{\delta}\left(\mathcal{S}^{\left(t\right)}, w^{\left(t\right)}\right)} \geq \tau$, it means that we have more  states to continue traversing to, and we can avoid the \knn{} search.
If $\abs{\hat{\delta}\left(\mathcal{S}^{\left(t\right)}, w^{\left(t\right)}\right)} < \tau$, it means that many of the entries at time $t$ were ``incorrect'' (their value was not equal to $w^{\left(t\right)}$), and the set of new states $\hat{\delta}\left(\mathcal{S}^{\left(t\right)}, w^{\left(t\right)}\right)$ is small (or even empty); in this case, we perform a new \knn{} search and restart an automaton traversal.

The minimal value is $\tau=1$, which means that as long as we have \emph{at least one} automaton state to continue traversing from, we use it without performing \knn{} search. The maximal value is $\tau=\infty$, which means that $\abs{\hat{\delta}\left(\mathcal{S}^{\left(t\right)}, w^{\left(t\right)}\right)}$ will always be lower than $\tau$, and thus we will perform \knn{} search at every step.

\section{Evaluation Details}
\label{app:hyperparams}
\paragraph{Implementation and Hyperparameters}
We used the exact hyperparameters of \cite{khandelwal2020generalization} including $k_{\text{neigh}}=1024$ (the number of retrieved nearest neighbors when performing a full search) and the same FAISS \cite{johnson2019billion} \knn{} library. Following \citet{he2021efficient}, we loaded the index to the GPU, and used half precision (fp16) datastore keys.

During the automaton traversal, when reaching automaton states -- there might be too many datastore entries in it, which can make the computation slow. We thus always computed \Cref{eq:pauto} over at most $\mathrm{max\_knns}=1024$ datastore entries (1024 datastore entries overall, in the union of all states $q\in\mathcal{S}^{\left(t\right)}$), preferring entries that were directly pointed by pointers from the previous state ($\rho\left(p\right)$), and otherwise choosing members from clusters randomly. We chose 1024 to match the number of nearest neighbors retrieved when performing a full search $k_{\text{neigh}}=1024$, although these numbers are not coupled to each other and denote different things.

\paragraph{Hardware} We ran all experiments on 32 CPU cores, and RTX 3090 or v100 GPUs. Since our main metric is the fraction of saved searches (FoSS), a different GPU will not change our results.
The experiments in \Cref{app:foss} (\Cref{fig:time}) were performed on the same machine using a RTX 3090 GPU.

\section{Fraction of Saved Searches (FoSS) vs. Wall-clock Saved Time}
\label{app:foss}
In our experiments in \Cref{sec:experiments}, we reported perplexity compared to FoSS (the fraction of saved searches).
The other alternative of measuring wall-clock time is difficult to reproduce, is brittle to %
temporary hardware and system load, and affected by the specific \knn{} retrieval library such as  FAISS as used in \citet{khandelwal2020generalization}, ScaNN \cite{guo2020accelerating} as used in \citet{borgeaud2021improving}, 
or SPTAG \cite{chen18sptag}, etc. 
Further, it depends on 
factors that are orthogonal to our contribution, 
such as whether the RAM is large enough to store the datastore, and the
 random-access reading latency of the hard-drive. %

FoSS, in contrast, does not depend on hardware, engineering factors, or temporary system load. FoSS is also completely reproducible while wall-clock time is not. Thus, FoSS serves as a good proxy to wall-clock time that enables reproducible experiments. 
Here, we perform an empirical analysis of FoSS with respect to the saved wall-clock time. 

\Cref{fig:time} shows a comparison between the fraction of saved wall-clock time and FoSS.
As shown, the saved wall-clock time and FoSS are identical up to an additive constant that depends on the hardware and the specific settings.
Searching for $k$-nearest neighbors using a CPU FAISS index results in a curve that is very close to the optimal $y=x$, meaning that almost the entire reduction in searches is translated directly into saving of wall-clock time.
Using a GPU index without clustering (only pointers) results in a penalty of 17\%, but the curve is almost parallel to the optimal $y=x$.
Using a GPU index with clustering results in a penalty of 24\%, and begins to be beneficial in terms of wall-clock time starting from FoSS=0.32.

The experiments in \Cref{fig:time} were performed in the setting that we load the datastore to memory, to prevent the hard drive's latency from being the bottleneck. Another option is to approximate the key vectors using the FAISS index, 
but currently FAISS's \texttt{reconstruct} API is implemented only for a CPU index (rather than a GPU index),\footnote{\url{https://github.com/facebookresearch/faiss/issues/314}} and for a single key ID at a time,\footnote{\url{https://github.com/facebookresearch/faiss/issues/1163}} and thus does not support batching.

We expect that as the datastore size increases to the scales of \citet{borgeaud2021improving}, and as the number of neighbors retrieved increases ($k_{\text{neigh}}$) -- the more pressure that will be put on the \knn{} search, the more of a bottleneck that it will become, and the larger relative benefit that saving \knn{} searches will provide to wall-clock time.

\input{appendix_analysis.tex}

\input{07_histogram.tex}

\input{ngram}

\input{fig_foss.tex}

%% file: appendix_analysis.tex
\section{Additional Results}
\label{app:analysis}
\paragraph{Comparison of Datasets}
\Cref{fig:ngram_types} and \Cref{fig:ngram_occ} show the overlap of n-grams between the training and validation set of \wiki{} and Law-MT. As shown, for all values of n, more n-grams from the validation set were seen 
in the training set in Law-MT compared to \wiki{}.
We see this as the explanation for the better scaling of \ourmodel{} on Law-MT, where the perplexity only gently increases as we increase FoSS,
compared \wiki{}.

\begin{figure*}[t]
	\begin{tikzpicture}[scale=1]
	\begin{axis}[
		xlabel={FoSS (fraction of saved searches)},
		ylabel={Perplexity},
		ylabel near ticks,
        legend style={at={(0,1)},anchor=north west,mark size=4pt,
        	}, %
        legend cell align={left},
        xmin=0, xmax=0.9,
        ymin=10.6, ymax=12,
        xtick={0.0, 0.1,0.2,...,1.0},
        ytick={10.0, 10.2,...,16},
        ylabel shift={-5pt},        
        grid = major,
        major grid style={dotted,gray},
                width=\textwidth
    ]

\addplot[color=gray, solid, mark options={solid, fill=white, draw=black}, mark=otimes, line width=2pt, mark size=4pt, visualization depends on=\thisrow{alignment} \as \alignment, nodes near coords, point meta=explicit symbolic,
    every node near coord/.style={anchor=\alignment, font=\footnotesize}] 
table [meta index=2]  {
x   y       label   alignment
0	10.85	{}		0
0.29	11.02	{}		0
0.38	11.20	{}		0
0.45	11.46	{}		0
0.51	11.79	{}		0
0.59	12.29	{}		0
0.64	12.69	{}		0
0.69	13.72	{}		0
0.76	16.83	{}		0
	};
    \addlegendentry{w/o clustering}

\addplot[color=red, solid, mark options={solid, fill=red, draw=black}, mark=triangle*, line width=2pt, mark size=4pt, visualization depends on=\thisrow{alignment} \as \alignment, nodes near coords, point meta=explicit symbolic,
    every node near coord/.style={anchor=\alignment, font=\footnotesize}] 
table [meta index=2]  {
x   y       label   alignment
0	10.64	{}		0
0.40	10.80	{}		0
0.49	10.87	{}		0
0.57	10.96	{}		0
0.64	11.03	{}		0
0.71	11.14	{}		0
0.75	11.26	{}		0
0.80	11.41	{}		0
0.84	11.80	{}		0
	};
    \addlegendentry{$k_{\mathrm{clust}}=$100K means}

\addplot[color=blue, mark options={solid, fill=blue, draw=black}, mark=square*, line width=2pt, mark size=4pt, visualization depends on=\thisrow{alignment} \as \alignment, nodes near coords, point meta=explicit symbolic,
    every node near coord/.style={anchor=\alignment, font=\footnotesize}] 
table [meta index=2]  {
x   y       label   alignment
0.0	10.61	{}		0
0.39	10.75	{}		0
0.48	10.81	{}		0
0.57	10.88	{}		0
0.64	10.96	{}		0
0.71	11.09	{}		0
0.75	11.21	{}		0
0.80	11.33	{}		0
0.84	11.79	{}		0
	};
    \addlegendentry{$k_{\mathrm{clust}}=$200K means}

\addplot[color=ao, mark options={solid, fill=ao, draw=black}, mark=*, line width=2pt, mark size=4pt, visualization depends on=\thisrow{alignment} \as \alignment, nodes near coords, point meta=explicit symbolic,
    every node near coord/.style={anchor=\alignment, font=\small
    }] 
table [meta index=2]  {
x   y       label   alignment
0	10.60	{}		0
0.38	10.72	{}		0
0.48	10.76	{}		0
0.56	10.83	{}		0
0.63	10.91	{}		0
0.70	11.06	{}		0
0.74	11.21	{}		0
0.79	11.39	{}		0
0.84	11.99	{}		0
	};
    \addlegendentry{$k_{\mathrm{clust}}=$400K means}

	\end{axis}
\end{tikzpicture}
\caption{Analysis of the number of clusters on the validation set of Law-MT.}
\label{fig:app:analysis_law}
\end{figure*}
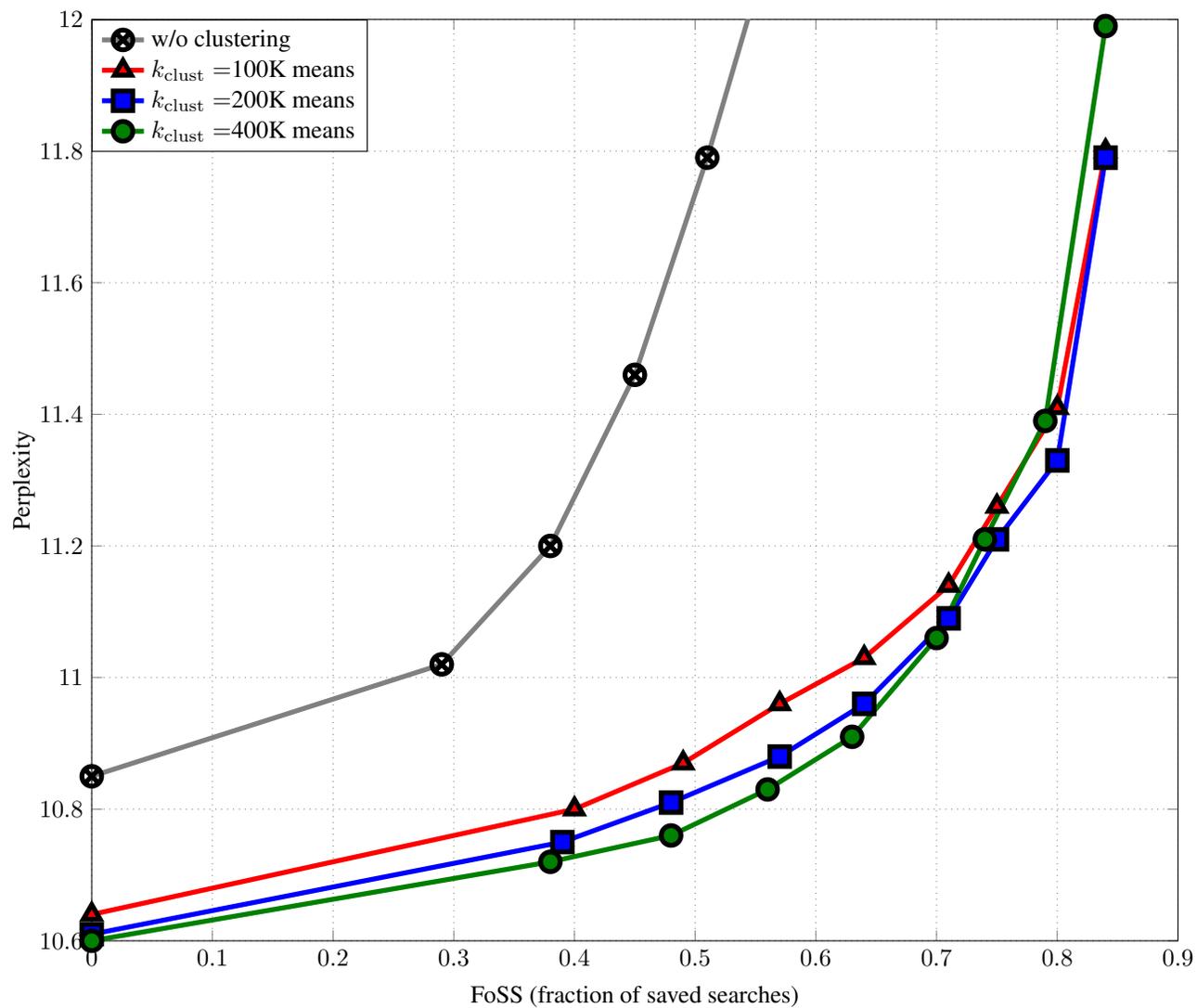

\paragraph{Ablation Study}
\Cref{fig:app:analysis_wiki} and \Cref{fig:app:analysis_law} show results for different clustering algorithms and granularities, on \wiki{} and Law-MT, respectively. These figures are similar to \Cref{fig:analysis_wiki} and \Cref{fig:analysis_law}, except that we include here more runs of $k$-means with more values of $k_{\text{clust}}$ and more runs of the greedy clustering of \citet{he2021efficient}.

\pagebreak

\definecolor{mypurple}{HTML}{AB30C4}

\begin{figure*}[t]
		\begin{tikzpicture}[scale=1]
	\begin{axis}[
		xlabel={FoSS (fraction of saved searches)},
		ylabel={Perplexity},
		ylabel near ticks,
        legend style={at={(0,1)},anchor=north west,mark size=4pt,
        	}, %
        legend cell align={left},
        xmin=0.0, xmax=0.9,
        ymin=16.09, ymax=17,
        xtick={0.1,0.2,...,1.0},
        ytick={16.1,16.2,...,17.0},
        ylabel shift={-5pt},        
        grid = major,
        major grid style={dotted,gray},
        width=\textwidth
    ]

\addplot[color=gray, solid, mark options={solid, fill=white, draw=black}, mark=otimes, line width=2pt, mark size=4pt, visualization depends on=\thisrow{alignment} \as \alignment, nodes near coords, point meta=explicit symbolic,
    every node near coord/.style={anchor=\alignment, font=\footnotesize}] 
table [meta index=2]  {
x   y       label   alignment
0	16.13	{}		0
0.20	16.20	{}		0
0.24	16.23	{}		0
0.29	16.26	{}		0
0.37	16.33	{}		0
0.43	16.39	{}		0	
0.50	16.48	{}		0
0.55	16.55	{}		0
0.59	16.63	{}		0
0.68	16.91	{}	0
	};
    \addlegendentry{w/o clustering}

\addplot[color=red, solid, mark options={solid, fill=red, draw=black}, mark=triangle*, line width=2pt, mark size=4pt, visualization depends on=\thisrow{alignment} \as \alignment, nodes near coords, point meta=explicit symbolic,
    every node near coord/.style={anchor=\alignment, font=\footnotesize}] 
table [meta index=2]  {
x   y       label   alignment
0	16.11	{}		0
0.29	16.21	{}	0
0.34	16.23	{}	0
0.39	16.26	{}	0
0.48	16.30	{}		0
0.55	16.33	{}	0
0.64	16.38	{}	0
0.70	16.43	{}	0
0.75	16.48	{}	0
0.83	16.69	{}	0
	};
    \addlegendentry{$k_{\mathrm{clust}}=$1M means}

\addplot[color=blue, mark options={solid, fill=blue, draw=black}, mark=square*, line width=2pt, mark size=4pt, visualization depends on=\thisrow{alignment} \as \alignment, nodes near coords, point meta=explicit symbolic,
    every node near coord/.style={anchor=\alignment, font=\footnotesize}] 
table [meta index=2]  {
x   y       label   alignment
0.0	16.11	{}	0
0.30	16.23	{}	0
0.35	16.25	{}	0
0.40	16.28	{}	0
0.48	16.32	{}	0
0.56	16.35	{}	0
0.65	16.41	{}	0
0.70	16.46	{}	0
0.75	16.51	{}	0
0.84	16.70	{}	0
	};
    \addlegendentry{$k_{\mathrm{clust}}=$500K means}

\addplot[color=ao, mark options={solid, fill=ao, draw=black}, mark=*, line width=2pt, mark size=4pt, visualization depends on=\thisrow{alignment} \as \alignment, nodes near coords, point meta=explicit symbolic,
    every node near coord/.style={anchor=\alignment, font=\small
    }] 
table [meta index=2]  {
x   y       label   alignment
0	16.1	{}	0
0.31	16.26	{}	0
0.41	16.32	{}	0
0.49	16.39	{}	0
0.56	16.42	{}	0
0.64	16.49	{}	0
0.70	16.56	{}	0
0.75	16.63	{}	0
0.84	16.82	{}	0
	};
    \addlegendentry{$k_{\mathrm{clust}}=$100K means}  
    
\addplot[color=mypurple, mark options={solid, fill=mypurple, draw=black}, mark=*, line width=2pt, mark size=4pt, visualization depends on=\thisrow{alignment} \as \alignment, nodes near coords, point meta=explicit symbolic,
    every node near coord/.style={anchor=\alignment, font=\small
    }] 
table [meta index=2]  {
x   y       label   alignment
0	16.15	{}	0
0.31	16.28	{}	0
0.35	16.31	{}	0
0.40	16.34	{}	0
0.48	16.40	{}	0
0.55	16.46	{}	0
0.63	16.54	{}	0
0.69	16.62	{}	0
0.74	16.70	{}	0
0.83	16.94	{}	0

	};
    \addlegendentry{$k_{\mathrm{clust}}=$50K means}     
  
\addplot[color=orange, mark options={solid, fill=orange, draw=black}, mark=diamond*, line width=2pt, mark size=4pt, visualization depends on=\thisrow{alignment} \as \alignment, nodes near coords, point meta=explicit symbolic,
    every node near coord/.style={anchor=\alignment, font=\small
    }] 
table [meta index=2]  {
x   y       label   alignment
0	16.09	{}	0
0.26	16.18	{}	0
0.30	16.20	{}	0
0.36	16.23	{}	0
0.45	16.28	{}	0
0.52	16.33	{}	0
0.59	16.40	{}	0
0.65	16.48	{}	0
0.69	16.57	{}	0
0.75	16.83	{}	0
	};
    \addlegendentry{greedy, 21M}

\addplot[color=cyan, mark options={solid, fill=cyan, draw=black}, mark=pentagon*, line width=2pt, mark size=4pt, visualization depends on=\thisrow{alignment} \as \alignment, nodes near coords, point meta=explicit symbolic,
    every node near coord/.style={anchor=\alignment, font=\small
    }] 
table [meta index=2]  {
x   y       label   alignment
0	16.10	{}	0
0.27	16.18	{}	0
0.31	16.21	{}	0
0.37	16.24	{}	0
0.46	16.28	{}	0
0.53	16.32	{}	0
0.61	16.40	{}	0
0.67	16.46	{}	0
0.71	16.53	{}	0
0.78	16.83	{}	0
	};
    \addlegendentry{greedy, 14M} 
    
	\end{axis}
\end{tikzpicture}
\caption{Analysis of the number of clusters on the validation set of \wiki{}. \jh{too big}}
\label{fig:app:analysis_wiki}
\end{figure*}

%% file: 07_histogram.tex
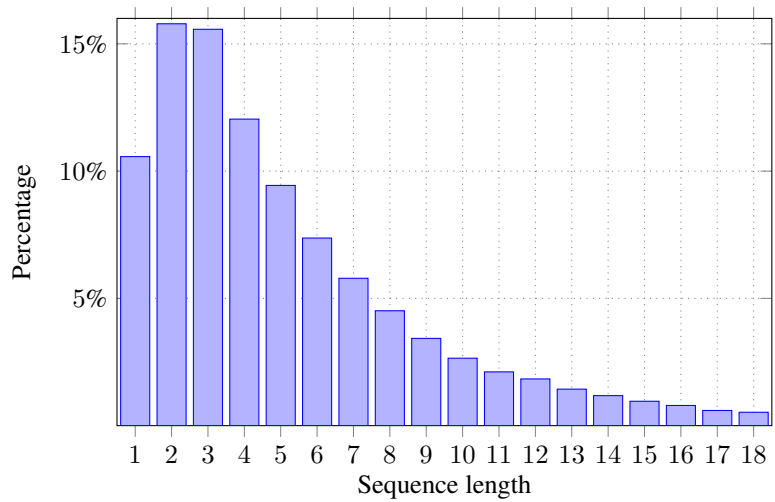
\begin{figure*}
\centering
\begin{tikzpicture}
 
\begin{axis} [ybar=12pt,
		bar width=11pt,
		xlabel={Sequence length},
		ylabel={Percentage},
        xmin=0.5	, xmax=18.5,
        ymin=0, ymax=0.16,
        xtick={1,...,18},
        ytick={0.05,0.10,...,1},
        grid = major,
        major grid style={dotted,gray},
	yticklabel={\pgfmathparse{\tick*100}\pgfmathprintnumber{\pgfmathresult}\%},
        height=7cm,
        width=0.6\textwidth,
        yticklabel style={
        /pgf/number format/fixed,
        /pgf/number format/precision=2,
},
scaled y ticks=false
] %
\addplot coordinates {

(1, 0.1056951565)
(2, 0.1578844161)
(3, 0.155727372)
(4, 0.1204302883)
(5, 0.0943776788)
(6, 0.07370367258)
(7, 0.05790402555)
(8, 0.04512984284)
(9, 0.03428859568)
(10, 0.0265008264)
(11, 0.02112222316)
(12, 0.01834888086)
(13, 0.01434294198)
(14, 0.01179370815)
(15, 0.009580637028)
(16, 0.007927837073)
(17, 0.005966887974)
(18, 0.00526654901)
(19, 0.004117993109)
(20, 0.00369778973)

};
\end{axis}
 
\end{tikzpicture}
\caption{Histogram of the lengths of sequences that were predicted consecutively, without \knn{} search, in \wiki{}.}
\label{fig:histo_lengths}
\end{figure*}

%% file: ngram.tex
\pgfplotstableread[row sep=\\,col sep=&]{
    n & wikitypes & wikioccur & lawtypes & lawoccur  \\
    1     & 1.0 & 1.0 & 0.9993212285762769 & 0.9999502134597912  \\
    2     & 0.8793132204143657 & 0.9323889005774131 & 0.9534328705724846 & 0.9798735406138757  \\
    3    & 0.632846433877315 & 0.6842587570832788 & 0.842751711351968 & 0.8896453865398738 \\
    4   & 0.36983238002474794 & 0.39537036604124687 & 0.7016280010184367 & 0.7513567338810058 \\
    5   & 0.20077170109763698 & 0.21171019534135646 & 0.5850482004253791 & 0.6253500790400677 \\
    6      & 0.11914188293237356 & 0.12403626315568003 & 0.48922554005071894 & 0.5205880156339465 \\
    7      & 0.0827863406543179 & 0.08520665793903123 & 0.41629606512184336 & 0.4407558161245752 \\
    8      & 0.06546936396623362 & 0.0667573721590245 & 0.3610808175414737 & 0.37964299940250945 \\
    9      & 0.055718764078690496 & 0.05631715276933579 & 0.3184967644329176 & 0.33230845833074 \\
    10      & 0.04908363769345817 & 0.04935547066772023 & 0.28340792434701256 & 0.29415938456942264 \\
    }\ngramdata

\begin{figure*}
\centering
\begin{tikzpicture}
 
\begin{axis} [ybar,
		xlabel={n-gram},
		ylabel={Percentage},
        legend cell align={left},
        symbolic x coords={1,2,3,4,5,6,7,8,9,10},
        ymin=0, ymax=1,
        xtick=data,
        grid = major,
        major grid style={dotted,gray},
	    yticklabel={\pgfmathparse{\tick*100}\pgfmathprintnumber{\pgfmathresult}\%},
        height=7cm,
        width=0.6\textwidth,
        yticklabel style={
        /pgf/number format/fixed,
        /pgf/number format/precision=2,
},
scaled y ticks=false
] %
\addplot table[x=n,y=wikitypes]{\ngramdata};
\addplot table[x=n,y=lawtypes]{\ngramdata};
\legend{\wiki{}, Law-MT}
\end{axis}
 
\end{tikzpicture}
\caption{The fraction of \emph{n-gram types} in the validation set that appeared verbatim in the training set in each dataset, for different values of n.}
\label{fig:ngram_types}
\end{figure*}

\begin{figure*}
\centering
\begin{tikzpicture}
 
\begin{axis} [ybar,
		xlabel={n-gram},
		ylabel={Percentage},
        legend cell align={left},
        symbolic x coords={1,2,3,4,5,6,7,8,9,10},
        ymin=0, ymax=1,
        xtick=data,
        grid = major,
        major grid style={dotted,gray},
	    yticklabel={\pgfmathparse{\tick*100}\pgfmathprintnumber{\pgfmathresult}\%},
        height=7cm,
        width=0.6\textwidth,
        yticklabel style={
        /pgf/number format/fixed,
        /pgf/number format/precision=2,
},
scaled y ticks=false
] %
\addplot table[x=n,y=wikioccur]{\ngramdata};
\addplot table[x=n,y=lawoccur]{\ngramdata};
\legend{\wiki{}, Law-MT}
\end{axis}
 
\end{tikzpicture}
\caption{The fraction of \emph{n-gram occurrences} in the validation set that appeared verbatim in the training set in each dataset, for different values of n.}
\label{fig:ngram_occ}
\end{figure*}

%% file: fig_foss.tex
\definecolor{ao}{rgb}{0.0, 0.5, 0.0}
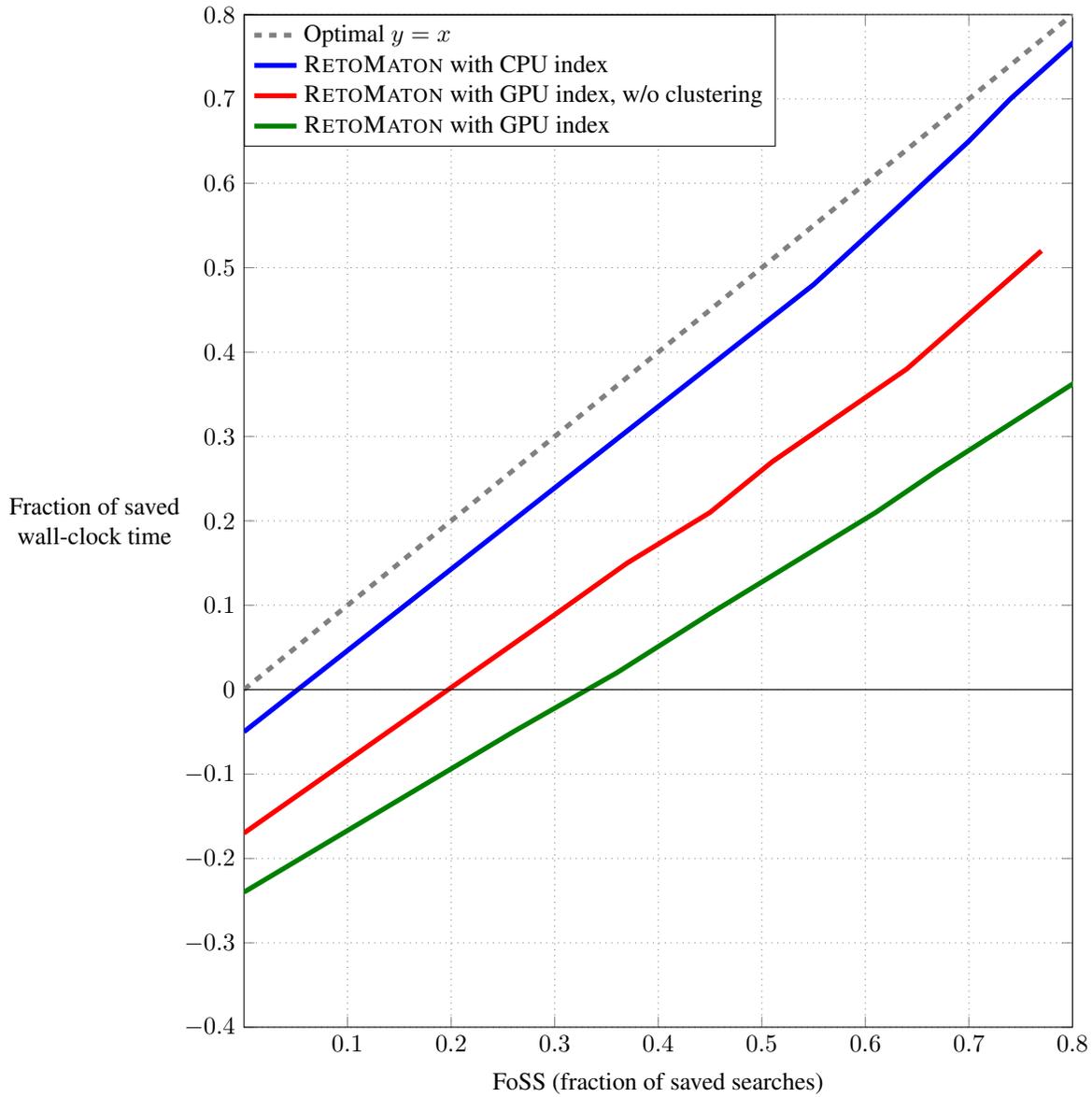
\begin{figure}[t]
	\begin{tikzpicture}[scale=1]
	\begin{axis}[
		xlabel={FoSS (fraction of saved searches)},
		ylabel={Fraction of saved \\ wall-clock time},
		ylabel near ticks,
        legend style={at={(0,1)},anchor=north west,mark size=2pt,
        	}, %
        legend cell align={left},
        xmin=0.0, xmax=0.8,
        ymin=-0.4, ymax=0.8,
        xtick={0.1,0.2,...,1.0},
        ytick={-0.4,-0.3,...,1.0},
        ylabel style={rotate=-90, align=center}, %
        ylabel shift={-5pt},        
        grid = major,
        major grid style={dotted,gray},
        width = 0.78\linewidth, height=16cm,
        yticklabel style={
        /pgf/number format/fixed,
        /pgf/number format/precision=1
		},
		scaled y ticks=false
    ]

\addplot[color=gray, dashed,
line width=2pt, 
visualization depends on=\thisrow{alignment} \as \alignment, nodes near coords, point meta=explicit symbolic,
    every node near coord/.style={anchor=\alignment, font=\small
    }] 
table [meta index=2]  {
x   y       label   alignment
0	0	{}		-10
1	1	{}	0
	};
    \addlegendentry{Optimal $y=x$}    

\addplot[color=blue, %
line width=2pt, visualization depends on=\thisrow{alignment} \as \alignment, nodes near coords, point meta=explicit symbolic,
    every node near coord/.style={anchor=\alignment, font=\footnotesize}] 
table [meta index=2]  {
x   y       label   alignment
0	-0.05	{}		-10
0.55	0.48	{}		-90
0.63	0.57	{}		-90
0.70	0.65	{}		-90
0.74	0.70	{}		-90
0.84	0.81	{}		-90
	};
    \addlegendentry{\ourmodel{} with CPU index}      

\addplot[color=red,
line width=2pt, 
visualization depends on=\thisrow{alignment} \as \alignment, nodes near coords, point meta=explicit symbolic,
    every node near coord/.style={anchor=\alignment, font=\small
    }] 
table [meta index=2]  {
x   y       label   alignment
0	-0.17	{}		-10
0.29	0.08	{}	0
0.37	0.15	{}	0
0.45	0.21	{}	0
0.51	0.27	{}	0
0.64	0.38	{}	0
0.77	0.52	{}	0
	};
    \addlegendentry{\ourmodel{} with GPU index, w/o clustering}   %

\addplot[color=ao, 
line width=2pt, 
visualization depends on=\thisrow{alignment} \as \alignment, nodes near coords, point meta=explicit symbolic,
    every node near coord/.style={anchor=\alignment, font=\small
    }] 
table [meta index=2]  {
x   y       label   alignment
0	-0.24	{}		0
0.26	-0.05	{}		0
0.36	0.02	{}		0
0.45	0.09	{}		0
0.53	0.15	{}		0
0.61	0.21	{}		0
0.67	0.26	{}		0
0.81	0.37	{}		0
	};
    \addlegendentry{\ourmodel{} with GPU index}

  \draw[thin] (axis cs:\pgfkeysvalueof{/pgfplots/xmin},0) -- (axis cs:\pgfkeysvalueof{/pgfplots/xmax},0);
  
	\end{axis}
\end{tikzpicture}
\caption{A comparison between the fraction of saved wall-clock time vs. FoSS, the fraction of saved searches. The fraction of saved wall-clock time was computed relatively to the baseline \knnlm{}.}
\label{fig:time}
\end{figure} 